\documentclass[twocolumn,10pt,cleanfoot]{asme2ej}
\usepackage[T1]{fontenc}
\usepackage{textcomp}  

\usepackage{amsmath}
\usepackage{bm} 
\usepackage{amssymb}   
\usepackage{wasysym}  
\usepackage{balance}
\usepackage{color} 
\usepackage{enumerate}
\usepackage{algorithm}
\usepackage{algorithmicx}
\usepackage{algpseudocode}

\makeatletter
\newcommand*{\algrule}[1][\algorithmicindent]{\makebox[#1][l]{\hspace*{.5em}\vrule height .75\baselineskip depth .25\baselineskip}}%
\newcount\ALG@printindent@tempcnta
\def\ALG@printindent{%
	\ifnum \theALG@nested>0
	\ifx\ALG@text\ALG@x@notext
	\addvspace{-3pt}
	\else
	\unskip
	\ALG@printindent@tempcnta=1
	\loop
	\algrule[\csname ALG@ind@\the\ALG@printindent@tempcnta\endcsname]%
	\advance \ALG@printindent@tempcnta 1
	\ifnum \ALG@printindent@tempcnta<\numexpr\theALG@nested+1\relax
	\repeat
	\fi
	\fi
}%
\usepackage{etoolbox}
\patchcmd{\ALG@doentity}{\noindent\hskip\ALG@tlm}{\ALG@printindent}{}{\errmessage{failed to patch}}
\makeatother
\newcommand*\myat{{\fontfamily{ptm}\selectfont @}}
\newcommand \AuthorEmail[1]{{\fontfamily{qcr}\fontsize{9}{10}\selectfont #1}}
\usepackage{tikz}
\newcommand{\realfield}[1]{\hbox{I \kern -.4em R}^{#1}}
\newcommand {\mb}[1]{\mathbf{#1}} 
\newcommand {\bs}[1]{\boldsymbol{#1}}
\newcommand{\uvec}[1]{\hat{\mathbf{#1}}}

\newcommand{\T}{^{\mathrm{T}}}  
\newcommand{\rmd}{\textrm{d}}  
\newcommand{\ith}{$i^{\,th}$ }
\newcommand*\circled[1]
{\tikz[baseline=(char.base)]{\node[circle,minimum size=7pt,draw=black,inner sep=0.5pt](char){\scriptsize #1};}}
\usepackage[normalem]{ulem}
\usepackage{color}  

\usepackage{array,url}
\newcommand{\Mark}[1]{\textsuperscript{#1}}

\usepackage{mathrsfs}  
\usepackage{amsbsy}  
\usepackage{arydshln} 
\usepackage{cases}
\usepackage{booktabs}

\newcommand{\msc}{\pmb{\mathscr{F}}}

\usepackage{color}  

\usepackage[textsize=scriptsize, bordercolor=white,backgroundcolor=gray!30,linecolor=black,colorinlistoftodos]{todonotes}  
\usepackage[normalem]{ulem} 
\usepackage{soul} 

\usepackage{xcolor}

\newcommand{\remind}[2]{{{#2}}}
\newcommand{\corrlab}[2]{{{#2}}}

\newcommand{\cusst}[1]{{}} 

\usepackage[titletoc,title,page]{appendix} 
\makeatletter
\def\@seccntformat#1{\@ifundefined{#1@cntformat}%
	{\csname the#1\endcsname\quad}
	{\csname #1@cntformat\endcsname}
}
\makeatother 

\usepackage{multirow}  
\usepackage{tabularx}  
\usepackage{colortbl} 
\usepackage{footnote}
\makeatletter
\newcommand{\thickhline}[1]{%
	\noalign {\ifnum 0=`}\fi \hrule height #1
	\futurelet \reserved@a \@xhline
}
\newcolumntype{"}{@{\hskip\tabcolsep\vrule width 2pt\hskip\tabcolsep}}
\makeatother 

\graphicspath{{figures_source/}}
\usepackage{hyperref}  
\usepackage[noadjust]{cite}
\renewcommand{\citedash}{--}

\usepackage{fancyhdr}
\begin{document}
\twocolumn[{%
	\centering
	{\huge Simplified Kinematics of Continuum Robot\\[0pt] Equilibrium Modulation via Moment \\[6pt] Coupling Effects and Model Calibration}\\[1.5em]
	\large Long~Wang\Mark{1},
	Giuseppe~Del~Giudice\Mark{1},
	and Nabil~Simaan\Mark{1}\\[1em]
	\normalsize
	\begin{tabular}{*{1}{>{\centering}p{0.4\textwidth}}}
		\Mark{1}Department of Mechanical Engineering
		\tabularnewline
		Vanderbilt University
		\tabularnewline
		\AuthorEmail{long.wang, giuseppe.del.giudice, nabil.simaan\myat Vanderbilt.edu}
	\end{tabular}\\[3em] 
}]

\begin{abstract}
	\it Recently, a new concept for continuum robots capable of producing macro-scale and micro-scale motion has been presented. These robots achieve their multi-scale motion capabilities by coupling direct-actuation of push-pull backbones for macro motion with indirect actuation whereby the equilibrium pose is altered to achieve micro-scale motion.  This paper presents a first attempt at explaining the micro-motion capabilities of these robots from a modeling perspective. This paper presents the macro and micro motion kinematics of a single segment continuum robot by using statics coupling effects among its sub-segments. Experimental observations of the micro-scale motion demonstrate a turning point behavior which could not be explained well using the current modeling methods. We present a simplistic modeling approach that introduces two calibration parameters to calibrate the moment coupling effects among the sub segments of the robot. It is shown that these two parameters can reproduce the turning point behavior at the micro-scale. The instantaneous macro and micro scale kinematics Jacobians and the calibration parameters identification Jacobian are derived. The modeling approach is verified against experimental data showing that our simplistic modeling approach can capture the experimental motion data with RMS position error of 5.82 $\mu m$  if one wishes to fit the entire motion profile with the turning point. If one chooses to exclude motions past the turning point, our model can fit the experimental data with an accuracy of 4.76 $\mu m$.
\end{abstract}

\section{Introduction}\label{ch:intro}
\par Current robotic manipulators for minimally invasive surgery (MIS) are capable of dexterous motion for surgical tasks requiring large workspace and position accuracy ranging from 0.5 to 1.5 mm. For example, the root mean square (RMS) localization accuracy of the da-Vinci Classic and da-Vinci S was evaluated experimentally as 1.02 mm and 1.05 mm respectively by Kwartowitz et al. \cite{Kwartowitz2006,Kwartowitz2007}. Despite recent increases in precision, current commercial surgical systems are unable to support micro-surgical precision (less than 0.1mm precision), and such precision can benefit micro-surgical tasks (e.g. micro-anastomosis and micro-vascular reconstruction \cite{Kanazawa2011,Slutsky2014,Chen2014,Lohmeyer2014}).
\begin{figure}[htbp]
	\centering
	\includegraphics[width= 0.7 \columnwidth]{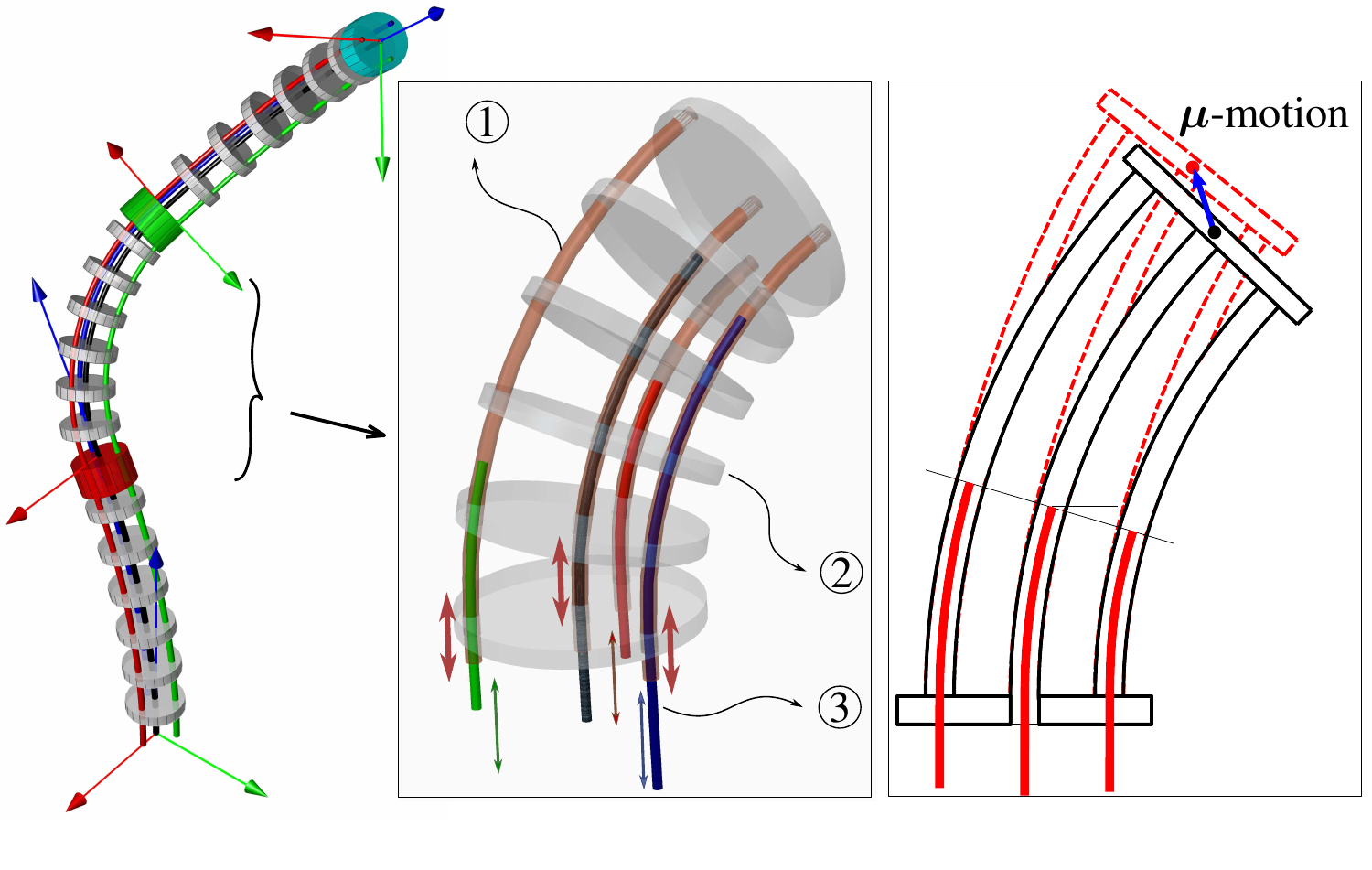}
	\vspace{-3mm}
	\caption{Continuum robots with equilibrium modulation (CREM): \protect\circled{1} secondary tubular backbones, \protect\circled{2} spacer disks, \protect\circled{3} equilibrium modulation backbones.
	}
	\label{fig:concept_CREM}
\end{figure}
\par This paper is motivated by a need for increased motion resolution at a micro-surgical scale and during deep surgical access minimally invasive surgery. In addition, the paper is equally motivated by the potential benefits of a new class of surgical devices capable of multi-scale motion. Such devices promise to provide a large workspace for traversal of deep passageways and for gross surgical manipulation while offering micro-scale motion suitable for cellular-level surgery. We refer to devices capable of macro and micro-scale motion as Multi Scale Motion (MSM) devices.  With the advent of new devices with integrated optical coherence tomography imaging and confocal endo-microscopy (e.g. \cite{yu2016calibration,zuo2017flexible}), the use of MSM can allow future image-based biopsy with imaging resolutions at cellular size \cite{fujimoto2003optical_biopsy, luo2005optical}. Such devices can in the future support surgical decisions on continued tumor excision, which can minimize the need for repeat follow-up surgeries due to incomplete resection of tumors.
\par To achieve MSM capabilities, this paper adopts the new design concept for \textit{Continuum Robot Equilibrium Modulation (CREM)}, which was first presented in \cite{delgiudice2017IROS_submitted}. CREM robots use a continuum structure that is primarily based on flexible elements to achieve large scale manipulation (i.e. robots without hinges and pin joints \cite{Robinson1999}). They also use fine adjustments to their static equilibrium pose in order to achieve micro-scale motion. The design concept for these robots is presented in Figure~\ref{fig:concept_CREM}. This design is modified from a multi-backbone continuum robot presented in \cite{Simaan2004}. Each segment of a multi-segment continuum robot (MBCR) includes superelastic NiTi backbones. A single central backbone is surrounded by secondary backbones that are radially constrained by spacer disks and equidistantly distributed circumferentially around the central backbone. Macro motion of the robot tip is achieved by pushing and pulling on the secondary backbones (designated by the thick arrows in Fig.~\ref{fig:concept_CREM}), which causes a deformation of the continuum robot body. We call this method of actuation \textit{direct actuation} where the robot actuators directly change the length of the secondary backbones. In addition, CREM robots use \textit{indirect actuation} whereby the equilibrium pose of the end effector is indirectly altered through a change of internal force distribution or by a change in material distribution altering cross sectional stiffness. For example, by inserting superelastic Ni-Ti beams (henceforth referred to as the \textit{Equilibrium Modulation Backbones} (EMBs)) inside the secondary backbones (see thin arrows in Fig.~\ref{fig:concept_CREM}), the static equilibrium of the robot is altered (modulated) by minute amounts.
\par Compared to prior designs, CREM robots possess a unique capability to allow MSM using a single design. Most prior works in the area of MSM rely on serial stacking of a macro-scale motion robot and a micro-scale robot, and such examples include Egeland's pioneering work \cite{Egeland1987} and followed by several other works such as \cite{comparetti2012accurate,Hodac1999,Entsfellner2012,Abiko2004,Cho2005,Kim2008,Nagatsu2013}. Other researchers investigated a variety of actuation methods and mechanisms to achieve micro motion capabilities, including a Steward/Gough parallel robot driven by hydraulic micro-actuators \cite{Portman2001}, twisted wire actuators for a planar parallel robot \cite{shoham2005twisting}, a micromanipulation tool using shape memory alloy \cite{Rul2007}, and a piezo-electrically actuated parallel platform \cite{Yun2008}. Although these works achieved micro-scale motion resolution, they are not suitable for MSM in confined spaces, in which continuum robots in general have advantages.
\par Within the context of continuum robots, the most relevant modeling works are \cite{Xu2010a} where a solution framework based on constraints of geometric compatibility and static equilibrium was derived using elliptic integrals for multi-backbone continuum robots and  \cite{rone2014continuum,rucker2011statics} where Cosserat rod theory was used for dynamics modeling of wire-actuated continuum robots. One could use these methods to model the tip micro motion, however, due to the formulation complexity and the solutions of equilibrium direct kinematics based on energy minimization or boundary value problem solution it is hard to obtain an updated differential kinematics model that accounts for the exact bending shape curvatures. Also, due to uncertainties in material properties and friction, using an exact modeling method does not add value since a model calibration step has to be carried out anyhow when attempting to control a physical robot.
\par Another work is Li \textit{et al.} \cite{li2016novel}, where the authors presented a constrained wire-driven flexible mechanism which used a constraint rod to selectively adjust its workspace by altering the length of its distal bending portion. The work showed that the constraint rod can change the workspace. The design however does not lend itself to easily allowing MSM and the work did not consider methods for achieving or modeling CREM.
\par Finally, our proposed design differs substantially from concentric tube robots \cite{dupont2010design, webster2009mechanics} in that the equilibrium modulation that creates the tip micro-motion is still governed by the strong geometric constraints employed by the parallel-backbone structure. Concentric tube continuum robots achieve their workspace through antagonistic bending of tube pairs and therefore they can theoretically be used for micro-scale equilibrium modulation. However to achieve micro-scale motion the designers are forced to use stiffness matched tube pairs with a very small difference in free curvature. The attainment of micro-scale motion by concentric tube robots therefore comes at the expense of sacrificing the macro-scale motion capabilities.
\par In contrast to the above-mentioned works, this paper takes a different approach. Instead of focusing on a high fidelity model, we present a simplified model that can be readily used to obtain the differential micro-motion kinematics Jacobian and is readily amenable to formulating a model calibration problem. This micro motion Jacobian is essential for control purpose, and an associated identification Jacobian is needed for calibration purposes. Therefore, the paper focuses on the derivation of the micro-motion kinematics and its associated identification Jacobian for calibration and error prorogation.
\par This work is built upon our previous work \cite{delgiudice2017IROS_submitted}, in which we presented the concept of CREM and provided a visual measurement solution to observe micro-motion. The work in \cite{delgiudice2017IROS_submitted} lacked a modeling approach that can explain the experimental observations and that can be used for control and identification purposes. The contribution of this work is in presenting a simplified kinematic modeling framework that captures the micro-motion achieved by the equilibrium modulation of continuum robots, and in developing a calibration approach to capture the model parameters. We put forth the concept of moment coupling effect as a simplified approach to describe the equilibrium modulation behavior, and thereby, both direct kinematics and instantaneous kinematics are formulated for control purposes. To account for errors potentially caused by the simplistic modeling assumptions, a modeling uncertainty term is introduced, and the identification Jacobian along with a calibration framework to capture the parameterization is developed. Using the multi-backbone continuum robot design in \cite{delgiudice2017IROS_submitted} as a validation platform, we validate the kinematic model and model calibration experimentally while augmenting these results with additional simulation validations.

\section[Equilibrium Shape Modeling]{Equilibrium Modulation Backbone Insertion to Create Micro Scale Motion}\label{ch:eqm}
This section presents the bending shape equilibrium modeling in the case where the Equilibrium Modulation Backbone (EMB) insertion is at a given depth $q_s$. To motivate the modeling approach taken we will first refer the readers to \cite{Simaan2004} where the simplified kinematics of multi-backbone continuum robots was presented. When the EMBs are not inserted and for proper design parameters (e.g. small spacing between the spacer disks) the continuum segment bends in a constant curvature \cite{Xu2008}. We use this assumption to create a simplistic equilibrium model which lends itself to fast real-time computation. Since we have to account for modeling uncertainties due to friction and material parameter uncertainties, we later lump the error of the simplified model in an uncertainty term $\lambda$ that will be used to produce an updated CREM model.
\subsection{Simplistic Equilibrium Model}
\begin{figure}[htbp]
	\centering
	\includegraphics[width= 0.9 \columnwidth]{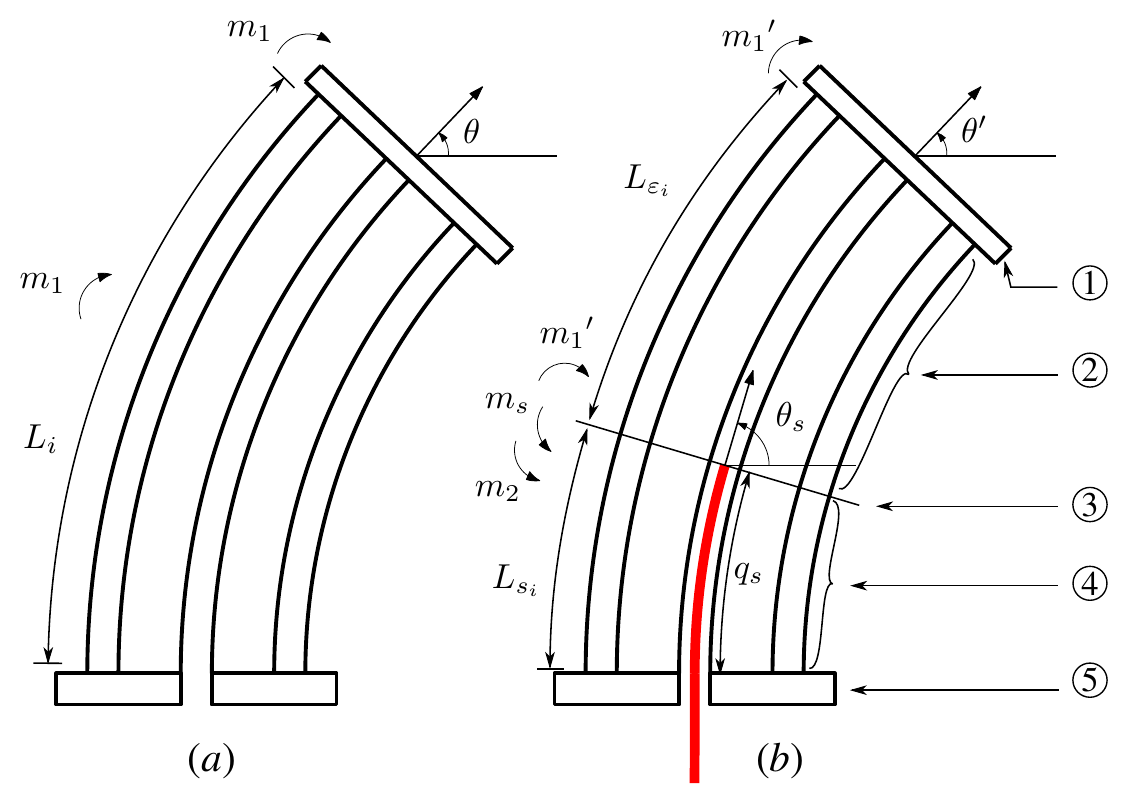}
	\vspace{-3mm}
	\caption{Example of a bent snake segment inserted with an equilibrium modulation backbone (EMB). For clarity, the spacer disks are not shown. \protect\circled{1} End-disk, \protect\circled{2} Empty subsegment, \protect\circled{3} Separation plane at EMB insertion depth $q_s$, \protect\circled{4} Inserted subsegment, \protect\circled{5} Base-disk.}
	\label{fig:SnakeSpring}
\end{figure}
%
\par Figure~\ref{fig:SnakeSpring} shows the free body diagram of a continuum segment with and without an inserted EMB. In Fig.~\ref{fig:SnakeSpring}(b), a separation plane is defined at the insertion depth $q_s$, dividing the segment into two subsegment - \textit{Inserted} and \textit{Empty}. Though not accurate, the two subsegments are both assumed to have constant but different curvatures. The angles $\theta'$ and $\theta_s$ denote the bending angles of the end-disk and at the insertion depth, respectively, when the EMB is inserted. The angle $\theta$ denotes the nominal bending angle when the EMB is not inserted. The angle $\theta_0=\pi/2$ denotes the angle at the base of the segment.
\par We first consider the resultant moment $m_1$ that the backbones apply on any imaginary cross section of the continuum segment when no EMB is inserted (Fig.~\ref{fig:SnakeSpring}(a)):
\begin{equation}\label{eqn:SnakeBend_M1_before_ins}
m_1 = \; E_p I_p \frac{\theta-\theta_0}{L} +
\sum\nolimits_{i} E_i I_i \frac{\theta-\theta_0}{L_i}
\end{equation}
\par \noindent Where $E_p$, $E_i $ and $I_p$, $I_i$ denote the Young's moduli and cross-sectional bending moments of inertia of the central backbone and the $i^{\textit{th}}$ secondary backbone, respectively.  Also $L$  and $L_i$ denote the lengths of the central backbone and the $i^{\textit{th}}$ secondary backbone.
\par We also consider the moment ${m_1}'$ along the empty subsegment in the case of EMB being inserted (Fig.~\ref{fig:SnakeSpring}(b)):
\begin{equation}\label{eqn:SnakeBend_M1_after_ins}
{m_1}' = \;E_p I_p \frac{\theta'-\theta_s}{L-q_s} +
\sum\nolimits_{i} E_i I_i \frac{\theta'-\theta_s}{L_{\varepsilon_i}}
\end{equation}
\par\noindent Where $L_{\varepsilon_i}$ denotes the $i^{\textit{th}}$ backbone length portion that belongs to the \textit{empty} subsegment (this is the arc-length from the separation plane until the end-disk).
\par We next use key definitions from \cite{Simaan2005}. The radial distance between the secondary backbones and the primary backbone is denoted $r$. When $r$ is projected onto the plane in which a segment bends, we obtain the projected radial distance $\Delta_i$:
\begin{equation}
\Delta_i=r\cos(\sigma_i),\quad \sigma_i= \delta + (i-1)\beta
\end{equation}
where $\sigma_i$ designates the angular coordinate of the \ith backbone relative to the bending plane. The angular coordinate of the first backbone relative to the bending plane is $\delta$ and the angular separation between secondary backbones is $\beta=\frac{2\pi}{n}$ where $n$ is the number of secondary backbones.
\par The length of the \ith backbone, $L_i$ is derived using the fixed radial offset between the backbones:
\begin{equation}\label{eqn:SnakeBend_li}
L_i =  \;L + \Delta_i (\theta - \theta_0)
\end{equation}
Using similar rationale, we calculate the empty length portion $L_{\varepsilon_i}$ and the inserted length portion of the \ith secondary backbone $L_{s_i}$:
\begin{align}
	&L_{s_i} =\;   q_s + \Delta_i (\theta_s - \theta_0) \label{eqn:SnakeBend_l_si}\\
	&L_{\varepsilon_i} = \;  (L - q_s) + \Delta_i (\theta' - \theta_s)= L_i-L_{s_i}\label{eqn:SnakeBend_l_ni}
\end{align}
\par In both Fig.~\ref{fig:SnakeSpring}(a) and (b), the static equilibrium at the end-disk is determined by the geometric constraints and the backbone loading forces at the end-disk. For example, coordinated pulling and pushing on all secondary backbones are assumed to form a force couple that generates only a moment at the end-disk.
\par We next use a simplifying assumption that the effect of EMB wire insertion on changes in the bending curvatures of the un-inserted subsegment backbones is negligible, hence:
\begin{equation}\label{eqn:SnakeBend:assumption}
m_1 = {m_1}'
\end{equation}
\par Next, we consider $m_2$ and $m_s$, the moments that the secondary backbones and the EMB apply on the separation plane as shown in Fig.~\ref{fig:SnakeSpring}(b):
\begin{align}
	m_2 = \; & -\left(E_p I_p \frac{\theta_s - \theta_0}{q_s} +
	\sum\nolimits_{i} E_i I_i \frac{\theta_s - \theta_0}{L_{s_i}}\right) \label{eqn:SnakeBend_M2}\\
	m_s = \;& -E_s I_s \frac{\theta_s - \theta_0}{q_s} \label{eqn:SnakeBend_Ms}
\end{align}
Where $E_s$ and $I_s$ denote the Young's modulus and cross-sectional bending moment of inertia of the EMB.
\par  Substituting equations~(\ref{eqn:SnakeBend_M1_before_ins},~\ref{eqn:SnakeBend_M1_after_ins}) into (\ref{eqn:SnakeBend:assumption}), results in one equation having two unknowns $\theta'$ and $\theta_s$ as illustrated in Fig.~\ref{fig:SnakeSpring}(b).  To obtain the second equation necessary for solving for these two unknowns, we use the moment balance on the separation plane:
\begin{equation}\label{eqn:SnakeBendM_govern}
{m_1}' + m_2 + m_s = 0
\end{equation}
\par\noindent To solve equations (\ref{eqn:SnakeBendM_govern}) and (\ref{eqn:SnakeBend:assumption}) for the unknowns $\theta'$ and $\theta_s$ we explicitly express the backbone moments using the beam equation $m=EI\kappa$ where $\kappa$ designates the radius of curvature and $EI$ designates the bending cross sectional stiffness of a beam.  In doing so, we note that the curvature of a beam bent in a circular shape satisfies $\kappa=\frac{\theta}{L}$ where $\theta$ is the deflection angle and $L$ is the beam length. Since the backbone lengths are a function of the unknowns, we will rewrite the moment equation for a beam as $m=\frac{EI}{L}\theta$ and, by defining the beam angular deflection stiffness $k_\theta\triangleq\frac{EI}{L}$ we obtain a simple equation for the moment $m=k_\theta\theta$.
\par Using the above definition for beam angular deflection stiffness, we rewrite the moment equations for each beam as:
\begin{eqnarray}\label{eqn:SnakeBendM_Re}
& m_1 = k_{\theta_0}\,(\theta - \theta_0),
& k_{\theta_0} =  \frac{E_p I_p}{L} +
\sum\nolimits_{i}  \frac{E_i I_i}{L_i} \label{eqn:SnakeBend_M1_k0}\\
& {m_1}' =  k_{\theta_1}\,(\theta' - \theta_s),
& k_{\theta_1} =  \frac{E_p I_p}{L-q_s} +
\sum\nolimits_{i}  \frac{E_i I_i}{L_{\varepsilon_i}} \label{eqn:SnakeBend_M1_k1}\\
& m_2 = -k_{\theta_2} \,(\theta_s - \theta_0),
& k_{\theta_2} = \frac{E_p I_p }{q_s} +
\sum\nolimits_{i} \frac{E_i I_i }{L_{s_i}}\label{eqn:SnakeBend_M2_k2}\\
& m_s = -k_{\theta_s} \,(\theta_s - \theta_0),
& k_{\theta_s} = \frac{E_s I_s}{q_s}\label{eqn:SnakeBend_Ms_ks}
\end{eqnarray}
\par \noindent Substituting Eq.~(\ref{eqn:SnakeBend_M1_k1})-(\ref{eqn:SnakeBend_Ms_ks}) in Eq.~(\ref{eqn:SnakeBendM_govern}) results in:
\begin{equation}\label{eqn:SnakeSpring_theta_s}
\theta_s = \dfrac{k_{\theta_1}\theta' + k_{\theta_2}\theta_0 + k_{\theta_s}\theta_0}{k_{\theta_1}+k_{\theta_2}+k_{\theta_s}}
\end{equation}
\par\noindent Substituting equations (\ref{eqn:SnakeBend_M1_k0}) and (\ref{eqn:SnakeBend_M1_k1}) in Eq. (\ref{eqn:SnakeBend:assumption}) results in:
\begin{equation}\label{eqn:SnakeSpring_theta_prime}
\theta' = \frac{k_0}{k_1}(\theta-\theta_0) + \theta_s = f_{\theta'}(\theta_s)
\end{equation}
As a final step in the solution, we substitute the result in Eq.~(\ref{eqn:SnakeSpring_theta_prime}) in Eq.~(\ref{eqn:SnakeSpring_theta_s}), thereby obtaining $\theta_s$ and subsequently $\theta'$.
\subsection{Updated CREM Model}
\par Equations (\ref{eqn:SnakeSpring_theta_s}) and (\ref{eqn:SnakeSpring_theta_prime}) present the solution to the simplistic modeling approach that is fundamentally based on  Eq.~(\ref{eqn:SnakeBend:assumption}) and Eq.~(\ref{eqn:SnakeBendM_govern}). In addition to the simplified assumption, the current model also neglects modeling uncertainties due to frictional effects and material property uncertainties. Prior works in \cite{goldman2014compliant, roy2017modeling} show that these uncertainties include friction and strain along the actuation lines, non-uniformly distributed load on backbones that causes shape deviations from constant-curvature bending, deviations in the cross section of the backbones during bending, and uncertainties in the properties of the NiTi backbones.
\par To account for the modeling uncertainties caused by friction, material uncertainty\footnote{Manufacturer-specified Young's modulus for superelastic NiTi is typically provided with a wide range of 40-70 GPa}, and our simplistic model, we introduce an uncertainty term $\lambda$ to Eq. (\ref{eqn:SnakeBendM_govern}):
\begin{equation}
{m_1}' + m_2 + m_s = \lambda(q_s,\theta,\mb{k}_\lambda)\label{eqn:SnakeBendM_govern_uncertainty}
\end{equation}
The uncertainty term $\lambda$ captures effects of EMB insertion offset, bending angle uncertainty and a fixed offset:
\begin{equation}\label{eqn:model_uncertainty_def}
\lambda(q_s,\theta,\mb{k}_\lambda) = k_{\lambda_0} + k_{\lambda_\theta}\,\theta + k_{\lambda_q} \,q_s; \quad
\mb{k}_\lambda\triangleq[k_{\lambda_0}, k_{\lambda_q}, k_{\lambda_\theta}]\T
\end{equation}
\par\noindent The solution in Eq.~(\ref{eqn:SnakeSpring_theta_s}) is also updated to:
\begin{equation}
\theta_s = \dfrac{k_1\theta' + k_2\theta_0 + k_s\theta_0 - \lambda}{k_1+k_2+k_s} \label{eqn:SnakeSpring_theta_s_uncertainty}
\end{equation}
\par Having obtained the solutions to the equilibrium tip bending angle $\theta'$ and the bending angle at the separation plane $\theta_s$, we introduce an \textit{equilibrium configuration space} variable vector $\bs{\phi}$ to combine them. With the purpose of preparing for kinematic derivations in later sections when we break a single continuum segment down to two subsegments, the vector $\bs{\phi}$ is defined as:
\begin{equation}\label{eqn:SnakeBend_Equilibrium_space_def}
\bs{\phi} \triangleq [\theta_s, \theta_{\varepsilon}]\T, \qquad \theta_{\varepsilon} \triangleq \theta' + \left(\tfrac{\pi}{2} -\theta_s\right)
\end{equation}
Where $\theta_{\varepsilon}$ represents the bending angle of the empty subsegment.
\par We define the \textit{configuration space} variable $\bs{\psi}$ as the \textit{nominal} bending angle $\theta$ and the bending plane angle $\delta$:
\begin{equation}\label{eqn:SnakeBend_Eqm_psi_def}
\bs{\psi} \triangleq [\theta, \delta]\T
\end{equation}
\par Finally, the solution to equilibrium modeling problem is presented as a mapping $\msc_\text{eqm}$ which is used in deriving the Jacobian matrices in the following sections:
\begin{equation}\label{eqn:SnakeBend_Equilibrium_func}
\bs{\phi}  =\msc_\text{eqm} (\bs{\psi}, q_s, \mb{k}_\lambda), \quad \bs{\phi}\in\realfield{2},\bs{\psi}\in\realfield{2}, \mb{k}_\lambda\in\realfield{3}
\end{equation}
\par \noindent Equation~(\ref{eqn:SnakeBend_Equilibrium_func}) provides the end disk equilibrium angle for a combination of any given EMB insertion length $q_s$, nominal bending angle $\theta$, and bending plane angle $\delta$. 

\section{Kinematic Modeling}\label{ch:kin}
Kinematic modeling of CREM includes the mapping of configuration space to task space and its differential kinematics. The differential kinematics include the instantaneous kinematics and the error propagation.
\par The instantaneous kinematics is derived for control purpose, and it includes two motion Jacobian matrices that both relate actuation speeds to the robot tip velocity. The \textit{macro} motion Jacobian $\mb{J}_M$ is associated with the joint velocities of push/pull on backbones (direct actuation) while the \textit{micro} motion Jacobian $\mb{J}_\mu$ is related to the EMB insertion velocity (indirect actuation).
\par The kinematic error propagation investigates how errors in parameters contribute to errors in kinematic measurements of task space (e.g. measured positions). In this work, we focus on the vector $\mb{k}_\lambda$ that parameterizes the modeling uncertainty. Other robot geometric kinematic parameters can be calibrated following \cite{Wang2014}. An \textit{identification} Jacobian $\mb{J}_\mb{k}$ is derived and used in section \ref{ch:calibration} to estimate $\mb{k}_\lambda$ with experimental measurements.

\subsection{Kinematic Modeling Using Mapping $\msc_\text{eqm}$}\label{ch:kin:kin}
\begin{figure}[htbp]
	\centering
	\includegraphics[width= 0.75 \columnwidth]{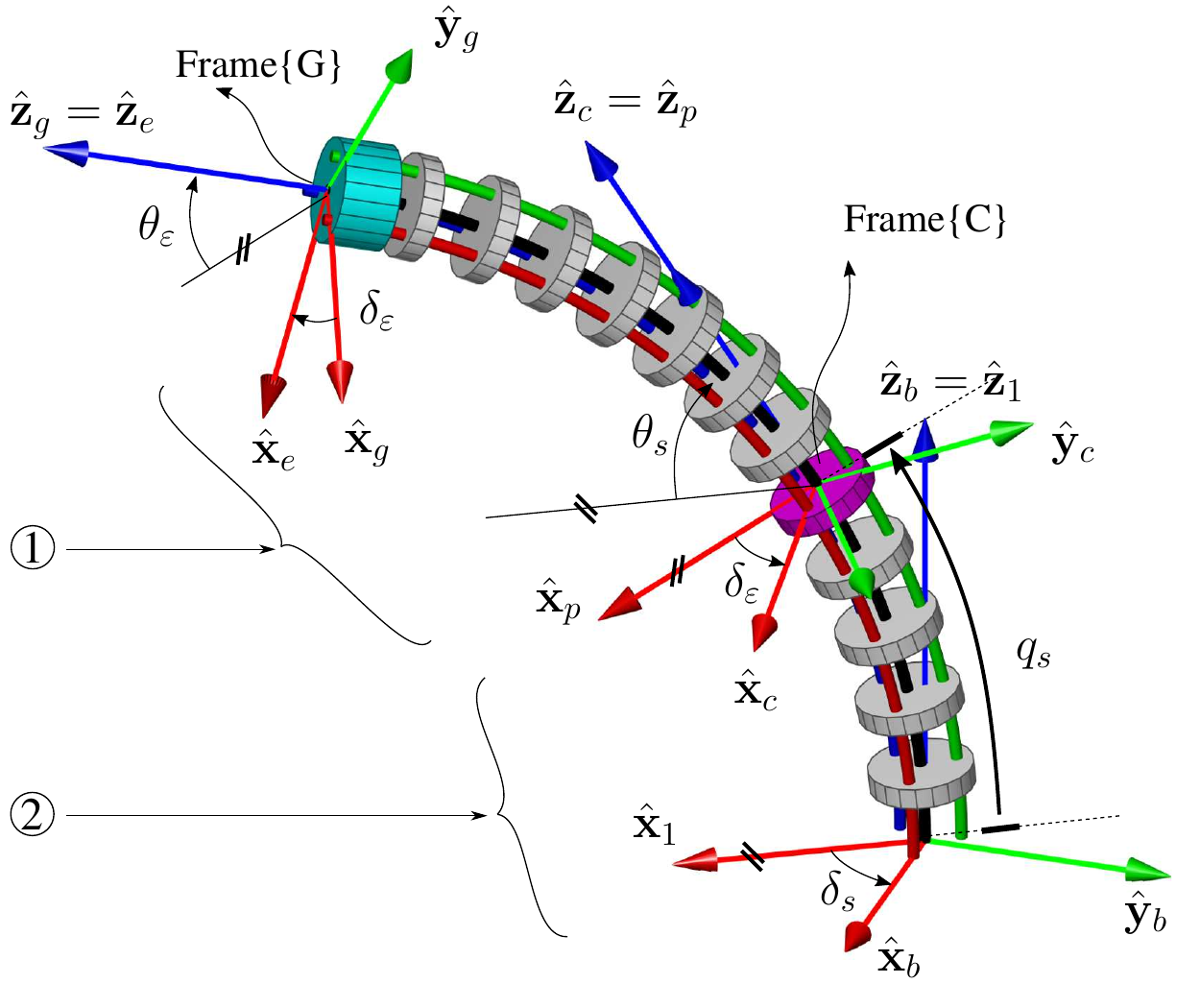}
	\caption{A single segment treated as two concatenated subsegments for a given micro-motion wire insertion depth. \protect\circled{1} Empty subsegment, \protect\circled{2} Inserted subsegment.}
	\label{fig:micro_insertion_kinematics}
\end{figure}
\par With the mapping $\msc_\text{eqm}$ in Eq.~(\ref{eqn:SnakeBend_Equilibrium_func}) derived as the result of static equilibrium, the kinematic mapping can be formulated by considering a single continuum segment as two concatenated subsegments - the inserted and the empty, divided at the insertion depth $q_s$. Figure~\ref{fig:micro_insertion_kinematics} illustrates our approach to analyzing the two concatenated subsegments. The bending angles of both subsegments were introduced in Eq.~(\ref{eqn:SnakeBend_Equilibrium_space_def}), denoted as $\theta_s$ and $\theta_{\varepsilon}$, for the inserted and the empty subsegment. Since the whole segment is assumed to bend in plane, both subsegments have the equal bending plane angles:
\begin{equation}\label{eqn:kin:delta_s_delta_r}
\delta_s \; = \delta_\varepsilon\;=  \delta
\end{equation}
\par \noindent The kinematic nomenclature used in this paper refers to Table~\ref{tab:nomenclature} (shown in Fig.~\ref{fig:micro_insertion_kinematics}).
\par Recalling the direct kinematics of a single segment \cite{goldman2014compliant} having length $L_x$ and an end disk angle $\theta_x$, the end disk pose (position and orientation) relative to the base are given by:
\begin{align}
	&  {}^{\text{base}\,}\mb{p}_{\text{\,end-disk / base}} = \frac{L_x}{\theta_x-\pi/2}
	\begin{bmatrix}
		\cos{\delta_x}\;(\sin{\theta_x}-1)\\
		-\sin{\delta_x}\;(\sin{\theta_x}-1)\\
		-\cos{\theta_x}
	\end{bmatrix} \label{eqn:segment_pos_kin}\\
	&{}^{\text{base}\,}\mb{R}_{\text{\,end-disk}} =
	e^{-\delta_x\,\left[\mb{z}^{\wedge}\right]}\;\;
	e^{(\frac{\pi}{2}-\theta_x)\,\left[\mb{y}^{\wedge}\right]}\;\;
	e^{\delta_x\,\left[\mb{z}^{\wedge}\right]}\label{eqn:segment_ori_kin}
\end{align}
Where $\delta_x$ designates the angle of the bending plane (analogous to $\delta$ in Fig. \ref{fig:micro_insertion_kinematics}), $\left[\mb{v}^{\wedge}\right]$ represents the cross-product matrix of vector $\mb{v}$ and the matrix exponential $e^{\;\alpha\left[\mb{v}^{\wedge}\right]}$ represents a rotation matrix about the axis $\mb{v}$ by an angle $\alpha$.
\par To obtain the pose of the end disk of the inserted segment is given by ${}^b\mb{p}_{c/b}$ and ${}^b\mb{R}_{c} $ we substitute $L_x=q_s, \theta_x=\theta_s, \delta_x=\delta$ in Eqs.~(\ref{eqn:segment_pos_kin},~\ref{eqn:segment_ori_kin}). Similarly, the pose of the end disk of the empty segment relative to its base is obtained by substituting $L_x=L-q_s, \theta_x=\theta_\varepsilon, \delta_x=\delta$ in Eqs.~(\ref{eqn:segment_pos_kin},~\ref{eqn:segment_ori_kin}) to result in ${}^c\mb{p}_{g/c} $ and  ${}^c\mb{R}_g$.
\par The pose of the free subsegment end disk relative to the segment base is given by:
\begin{align}
	&{}^b\mb{p}_{g/b}  = {}^b\mb{p}_{c/b} + {}^b\mb{R}_c \,{}^{c}\mb{p}_{g/c} \;\triangleq  \msc_g(\bs{\phi},\delta, q_s)
	\label{eqn:KinMicro:g_in_b}\\
	& \label{eqn:KinMicro:g2b}
	^b\mb{R}_g = {}^b\mb{R}_c \,{}^c\mb{R}_g\; = \mb{R}_{g}(\bs{\phi},\delta)
\end{align}
Casting the above two equations in a homogeneous transform format yields:
\begin{equation}\label{eqn:Kin:Homog}
{}^b\mb{T}_g = \left[\begin{array}{cc}
^b\mb{R}_g &  ^b\mb{p}_{g/b}  \\
\mb{0} & 1
\end{array}
\right] \triangleq \msc_T(\bs{\phi},\delta, q_s)
\end{equation}
\par \noindent With $\bs{\phi}$ expressed using mapping $\msc_\text{eqm}(\bs{\psi},q_s,\mb{k}_\lambda)$, the forward kinematics is determined, which can be also written as:
\begin{equation}\label{eqn:Kin:Homog2}
{}^b\mb{T}_g \triangleq \msc_T(\bs{\psi},q_s,\mb{k}_\lambda)
\end{equation}
%


\subsection{Differential Kinematics}\label{ch:kin:differential}
The total differential of a homogenous transformation $\mb{T}\in\text{SE}(3)$, may be represented as:
\begin{align}
	&\rmd \bm{\xi} \triangleq \left[(\rmd\mb{x})\T, (\rmd\boldsymbol{\mu})\T\right]\T, \quad \rmd\bs{\xi}\in\realfield{6\times1}
	\label{eqn:Kin:Diff_Paul}\\
	&\rmd\mb{x}  \triangleq \rmd ({}^b\mb{p}_{g/b}), \quad\rmd\boldsymbol{\mu} \triangleq [\rmd \mu_x, \rmd \mu_y, \rmd \mu_z]\T
\end{align}
Where $\rmd\mb{x}$ and $\rmd\bs{\mu}$ represent translational and rotational differentials in the base frame\footnote{A differential rotation is a sequence of rotations of small angles.}. The vector $\bs{\mu}\in\realfield{3\times 1}$ represents a chosen orientation parametrization (e.g. Euler angles).
\par The total differential of ${}^b\mb{T}_g$ is obtained by considering differentials on all variables, i.e. $\rmd \bs{\phi}$, $\rmd \delta$, and $\rmd q_s$:
\begin{equation}
\rmd \bs{\xi} =
\dfrac{\partial \bs{\xi}}{\partial \bs{\phi}} \;\rmd \bs{\phi} \;+\;
\dfrac{\partial \bs{\xi}}{\partial \delta} \;\rmd \delta \;+\;
\dfrac{\partial \bs{\xi}}{\partial q_s} \;\rmd q_s \vspace{1mm}\label{eqn:Kin:Diff_full}
\end{equation}
Using the nomenclature of a Jacobian $\mb{J}_{ab}$ such that $\delta a = \mb{J}_{ab} \delta b$, we define the following Jacobian matrices:
\begin{equation}
\tfrac{\partial \bs{\xi}}{\partial \bs{\phi}} \triangleq \mb{J}_{\bs{\xi}\bs{\phi}}\in\realfield{6\times2},\quad
\tfrac{\partial \bs{\xi}}{\partial \delta} \triangleq \mb{J}_{\bs{\xi}\delta}\in\realfield{6},\quad
\tfrac{\partial \bs{\xi}}{\partial q_s} \triangleq \mb{J}_{\bs{\xi}q_s}\in\realfield{6} \label{eqn:Kin:Diff_full_Jac}
\end{equation}
%
\par \noindent The Jacobian matrices $\mb{J}_{\bs{\xi}\bs{\phi}}$, $\mb{J}_{\bs{\xi}\delta}$, and $\mb{J}_{\bs{\xi}q_s}$, respectively, relate the differential on equilibrium configuration space variable $\rmd \bs{\phi}=[\rmd \theta_s,\rmd \theta_{\varepsilon}]\T$, the differential on bending plane angle $\rmd \delta$, and differential on EMB insertion depth $\rmd q_s$, to the corresponding differential contributions on the pose vector $\rmd \bs{\xi}$.
\par Both $\mb{J}_{\bs{\xi}\bs{\phi}}$ and $\mb{J}_{\bs{\xi}\delta}$ can be obtained by treating the inserted and empty subsegments as a concatenated two-segment robot, which is explained in section \ref{ch:kin:2seg}.
\par The third Jacobian, $\mb{J}_{\bs{\xi}q_s}$, defined as the partial derivative, ${\partial \bs{\xi}}/{\partial q_s}$, is derived with the other variables ($\bs{\phi}$ and $\delta$) held constant. The end-effector orientation, given by ${}^b\mb{R}_g$ in Eq.~(\ref{eqn:KinMicro:g2b}) is not a function of $q_s$. Therefore, by considering only the translational differential due to $\rmd q_s$, we have:
\begin{equation}\label{eqn:Kin:J_x_qs}
\mb{J}_{\bs{\xi}q_s} =
\left[\begin{array}{c}
\dfrac{\partial\;{}^b\mb{p}_{c/b}}{\partial q_s}  +
{}^b\mb{R}_c\;\dfrac{\partial\;{}^c\mb{p}_{g/c}}{\partial q_s} \vspace{1mm}\\\hdashline[2pt/2pt]
\mb{0}_{3\times1}
\end{array}\right]
\end{equation}
%
%
\begin{table}[htbp]
	\centering
	\footnotesize
	\caption{Nomenclature for Kinematic Modeling}
	\begin{tabular}{m{.10\columnwidth} m{.82\columnwidth}}
		Symbol & Description \\
		\thickhline{1.5pt}
		Frame \{F\}
		& designates a right-handed frame with unit vectors $\uvec{x}_f,\uvec{y}_f,\uvec{z}_f$ and point $\mb{f}$ as its origin.\\
		\hdashline
		Frame \{B\}
		& the base disk frame with $\mb{b}$ located at the center of the base disk, $\uvec{x}_b$ passing through the first secondary backbone and $\uvec{z}_b$ perpendicular to the base disk.\\
		\hdashline
		Frame \{1\}
		& Frame of the bending plane having $\uvec{z}_1=\uvec{z}_b$ and $\uvec{x}_1$ passing through with the project point of the end disk center. The angle $\delta$ is defined as from $\uvec{x}_1$ to $\uvec{x}_b$ about $\uvec{z}_b$ according to right hand rule.    \\
		\hdashline
		\parbox{0.8cm}{Frames \{E\} \& \{G\}}
		& Frame \{E\} is defined with $\uvec{z}_e$ as the normal to the \textit{end} disk and $\uvec{x}_e$ is the intersection of the bending plane and the end disk top surface. Frame \{G\} is the \textit{gripper} frame that has the same $\uvec{z}$ as \{E\}, i.e. $\uvec{z}_g=\uvec{z}_e$, but $\uvec{x}_g$ is passing through the first secondary backbone. It can be obtained by a rotation angle $\left(-\sigma_{1e}\right)$ about $\uvec{z}_e$.\\
		\hdashline
		Frames \newline \{P\} \& \{C\}
		& These frames are defined in a manner similar to the definition of frames \{E\} and \{G\} but for a specific arc insertion length $q_s$ as opposed to the full length of the robot segment $L$. The $\uvec{x}_{c}-\uvec{y}_{c}$ plane is the insertion plane as in shown in the planar case in Fig.~\ref{fig:SnakeSpring}.\\
		\hdashline
		Frame \newline \{I\}
		& designates the microscope image frame having the origin at the corner of the image and having its $x$-$y$ axes aligned with the width and height directions \big(Fig.~\ref{fig:exp_setup}(a, c)\big).\\
		\hdashline
		Frame \newline \{M\}
		& designates the marker frame that is determined by segmentation of the microscope image \big(Fig.~\ref{fig:exp_setup}(c)\big).\\
		\hdashline
		Vector \newline $^x\mb{p}_{a/b}$
		& designates the position of point $a$ relative to point $b$ that is expressed in frame \{X\}.\\
		\thickhline{1pt}
	\end{tabular}
	\label{tab:nomenclature}
\end{table}
%
Where $\tfrac{\partial\;{}^b\mb{p}_{c/b}}{\partial q_s}$ and $\tfrac{\partial\;{}^c\mb{p}_{g/c}}{\partial q_s}$ are derived from Eq.~(\ref{eqn:segment_pos_kin}). It is important to note that $\mb{J}_{\bs{\xi}q_s}$ differs from the micro motion Jacobian $\mb{J}_{\mu}$ derived later in that $\mb{J}_{\bs{\xi}q_s}$ is a contributing part of $\mb{J}_{\mu}$ - the length `tangential' contribution, while $\rmd q_s$ also propagates to $\rmd\bs{\phi}$ that also causes change on $\rmd\bs{\xi}$.
\par Having derived the above three Jacobian matrices, $\mb{J}_{\bs{\xi}\bs{\phi}}$, $\mb{J}_{\bs{\xi}\delta}$, and $\mb{J}_{\bs{\xi}q_s}$, we obtain the pose total differential $\rmd\bs{\xi}$ expressed using differentials $\rmd\bs{\phi}$, $\rmd \delta$, and $\rmd q_s$. Further, the differential $\rmd\bs{\phi}$ is a result of multiple other differentials, which can be seen from mapping $\msc_\text{eqm}(\bs{\psi},q_s,\mb{k}_\lambda)$. To fully investigate and decouple the contributions of direct (macro) and indirect (micro) actuation, we express $\rmd\bs{\phi}$ using differentials on $(\bs{\psi}, q_s, \mb{k}_\lambda)$. This differentiation is also motivated by Eq.~(\ref{eqn:Kin:Homog2}), where the variables are decoupled as $\bs{\psi}$ for macro motion, $q_s$ for micro motion, and $\mb{k}_\lambda$ for micro motion parameters. Such differentiation is derived as:
\begin{gather}
	\rmd \bs{\phi} = \dfrac{\partial \bs{\phi}}{\partial \bs{\psi}} \rmd \bs{\psi}+ \dfrac{\partial \bs{\phi}}{\partial q_s} \rmd q_s + \dfrac{\partial \bs{\phi}}{\partial \mb{k}_\lambda} \rmd \mb{k}_\lambda \label{eqn:Kin:d_phi_full}\\
	\dfrac{\partial \bs{\phi}}{\partial \bs{\psi}} = \left[\tfrac{\partial \bs{\phi}}{\partial \theta},\,\tfrac{\partial \bs{\phi}}{\partial \delta}\right], \quad \dfrac{\partial \bs{\phi}}{\partial \mb{k}_\lambda}  = \left[\tfrac{\partial \bs{\phi}}{\partial {k}_{\lambda_0}}, \tfrac{\partial \bs{\phi}}{\partial {k}_{\lambda_\theta}}, \tfrac{\partial \bs{\phi}}{\partial {k}_{\lambda_q}}\right]
\end{gather}
Where the gradient terms are derived in Appendix~\ref{app:d_phi} as:
\begin{align}
	& \dfrac{\partial \bs{\phi}}{\partial \theta} =
	\left(\mb{A}\,\mb{S}_0 - \bs{\Gamma}_{\theta_s}\mb{S}_1\right)^{-1}\;\bs{\Gamma}_\theta
	\;\triangleq\mb{J}_{\bs{\phi}\theta} \label{eqn:Kin:d_phi_d_theta}\\
	& \dfrac{\partial \bs{\phi}}{\partial \delta} =
	\left(\mb{A}\,\mb{S}_0 - \bs{\Gamma}_{\theta_s}\mb{S}_1\right)^{-1}\;\bs{\Gamma}_\delta
	\;\triangleq\mb{J}_{\bs{\phi}\delta} \label{eqn:Kin:d_phi_d_delta}\\
	&  \dfrac{\partial \bs{\phi}}{\partial q_s}  =
	\left(\mb{A}\,\mb{S}_0 - \bs{\Gamma}_{\theta_s}\mb{S}_1\right)^{-1}\; \bs{\Gamma}_{q_s}
	\;\triangleq\mb{J}_{\bs{\phi}q_s}\label{eqn:Kin:d_phi_d_q_s}\\
	& \dfrac{\partial \bs{\phi}}{\partial k_{\lambda_i}} =
	\left(\mb{A}\,\mb{S}_0 - \bs{\Gamma}_{\theta_s}\mb{S}_1\right)^{-1}\;\mb{B}'_{k_{\lambda_i}}
	\triangleq\mb{J}_{\bs{\phi}k_{\lambda_i}}, \; \dfrac{\partial \bs{\phi}}{\partial \mb{k}_\lambda} \triangleq \mb{J}_{\bs{\phi}\mb{k}_\lambda}
	\label{eqn:Kin:d_phi_d_k_lambda}\\
	& \mb{C}_{\bs{\phi}} \triangleq \mb{S}_0\bs{\phi} - \mb{C}_0, \quad
	\bs{\Gamma}_x = \mb{B}'_x - \mb{A}'_x \mb{C}_{\bs{\phi}}
\end{align}
Matrices $\mb{A}'_x, \mb{B}'_x$ are partial derivative matrices with respect to variable `$x$', and $\mb{A}, \mb{S}_0, \mb{B}, \mb{C}_0, \mb{S}_1$ are defined as:
\begin{align}
	& \mb{A} = \left[\begin{array}{cc}
		k_1+k_2+k_s & -k_1 \\
		k_1 & -k_1
	\end{array}  \right]
	,\; \mb{S}_0 = \left[\begin{array}{cc}
		1 & 0 \\
		1 & 1
	\end{array}
	\right] \label{eqn:Kin:matrix_A_S}\\
	& \mb{B} = \left[\begin{array}{c}
		(k_2 + k_s)\,\theta_0 - \lambda \\
		k_0(\theta_0-\theta)
	\end{array}
	\right],\;
	\mb{C}_0 = \left[ \begin{array}{c}
		0 \\
		\theta_0
	\end{array}
	\right], \;
	\mb{S}_1 = \left[ \begin{array}{cc}
		1 &
		0
	\end{array}
	\right]
	\label{eqn:Kin:matrix_B_C0}
\end{align}
\par \noindent Using Eq.~(\ref{eqn:Kin:d_phi_d_theta}-\ref{eqn:Kin:matrix_B_C0}), $\rmd \bs{\phi}$ is fully expressed as Eq.~(\ref{eqn:Kin:d_phi_full}). Substituting $\rmd \bs{\phi}$ into the original differentiation in Eq.~(\ref{eqn:Kin:Diff_full}), we obtain the full differential kinematics that relates differentials on $\{\bs{\psi}, q_s, \mb{k}_\lambda\}$ to the pose total differential $\rmd \bs{\xi}$:
\begin{equation}\label{eqn:Kin:diff_full_DT_result}
\begin{array}{ll}
\rmd \bs{\xi} =
& \dfrac{\partial \bs{\xi}}{\partial \bs{\phi}} \, \dfrac{\partial \bs{\phi}}{\partial \theta}\,\rmd \theta \;+\;
\dfrac{\partial \bs{\xi}}{\partial \bs{\phi}} \,\dfrac{\partial \bs{\phi}}{\partial \delta} \,\rmd \delta\;+\;
\dfrac{\partial \bs{\xi}}{\partial \bs{\phi}} \,\dfrac{\partial \bs{\phi}}{\partial q_s} \,\rmd q_s \;+
\vspace{5pt}\\
&
\dfrac{\partial \bs{\xi}}{\partial \bs{\phi}} \,\dfrac{\partial \bs{\phi}}{\partial \mb{k}_\lambda} \,\rmd \mb{k}_\lambda \;+
\dfrac{\partial \bs{\xi}}{\partial \delta} \,\rmd \delta \;+
\dfrac{\partial \bs{\xi}}{\partial q_s} \,\rmd q_s
\end{array}
\end{equation}
Rewriting Eq. (\ref{eqn:Kin:diff_full_DT_result}) using the Jacobian definitions:
\begin{equation}\label{eqn:Kin:diff_full_DT_result}
\begin{array}{ll}
\rmd \bs{\xi} =
& \mb{J}_{\bs{\xi}\bs{\phi}} \,\mb{J}_{\bs{\phi}\theta} \,\rmd \theta \;+\;
\mb{J}_{\bs{\xi}\bs{\phi}} \,\mb{J}_{\bs{\phi}\delta} \,\rmd \delta\;+\;
\mb{J}_{\bs{\xi}\delta} \,\rmd \delta \;+ \\
&  \mb{J}_{\bs{\xi}\bs{\phi}} \,\mb{J}_{\bs{\phi}q_s} \,\rmd q_s \;+
\mb{J}_{\bs{\xi}q_s} \,\rmd q_s \;+
\mb{J}_{\bs{\xi}\bs{\phi}} \,\mb{J}_{\bs{\phi}\mb{k}_\lambda} \,\rmd \mb{k}_\lambda
\end{array}
\end{equation}
Collecting like terms of $\rmd\bs{\psi}$, $\rmd q_s$, and $\rmd \mb{k}_\lambda$, we obtain a differentiation that decouples differentials of the macro motion, the micro motion, and the parameters:
\begin{equation}\label{eqn:Kin:diff_full_DT_Jacobian_defs}
\begin{array}{ll}
\rmd \bs{\xi} =&\underbrace{\left[\begin{array}{c;{2pt/2pt}c}
	\mb{J}_{\bs{\xi}\bs{\phi}} \,\mb{J}_{\bs{\phi}\theta}&\mb{J}_{\bs{\xi}\bs{\phi}} \,\mb{J}_{\bs{\phi}\delta} + \mb{J}_{\bs{\xi}\delta}
	\end{array}\right]}_{\triangleq\;\; \mb{J}_{M_\psi}} \rmd\bs{\psi}        \;+\;\\
&\underbrace{\left[\mb{J}_{\bs{\xi}\bs{\phi}} \,\mb{J}_{\bs{\phi}q_s} + \mb{J}_{\bs{\xi}q_s} \right]}_{\triangleq\;\;\mb{J}_\mu}\, \rmd q_s
\;+\;
\underbrace{\mb{J}_{\bs{\xi}\bs{\phi}} \,\mb{J}_{\bs{\phi}\mb{k}_\lambda}}_{\triangleq\;\;\mb{J}_\mb{k}} \,\rmd \mb{k}_\lambda
\end{array}
\end{equation}
\par\noindent The above result completes the mapping from configuration to task space. It clearly delineates the effects of EDM insertion and direct actuation to achieving macro and micro motion. For control purposes, a complete mapping from joint to task space is needed. We therefore consider next the mapping from direct (macro) actuation joint space $\mb{q}$ to task space $\bs{\xi}$. Since three secondary backbones are used in our experimental setup as illustrated in Figure~\ref{fig:micro_insertion_kinematics}, we will define $\mb{q}\triangleq[q_1, q_2, q_3]\T$ where:
\begin{equation}
q_i\triangleq L_i-L
\end{equation}
When obtaining this mapping, we consider the nominal segment kinematics for multi-backbone continuum robots as in \cite{Simaan2004}.
\par The Jacobian that relates the differential $\rmd \mb{q}$ to the differential $\rmd\bs{\psi}$ was reported in \cite{Xu2008} as:
\begin{equation}\label{eqn:kin:J_q_psi}
\rmd \mb{q} \triangleq \mb{J}_{\mb{q}\bs{\psi}}\;\rmd\bs{\psi}, \quad
\mb{J}_{\mb{q}\bs{\psi}} =
r
\begin{bmatrix}
c_\delta                        &  (\theta_0-\theta) \;s_\delta \\
c_{(\delta+\beta)}     & (\theta_0-\theta)  \;s_{(\delta+\beta)}\\
c_{(\delta+2\beta)}  & (\theta_0-\theta)  \;s_{(\delta+2\beta)}
\end{bmatrix}
\end{equation}
Where $r$ denotes the constant radial distance between the central and surrounding backbones, and $\beta=2\pi/3$ denotes the backbone separation angle. Using Eq.~(\ref{eqn:kin:J_q_psi}), we substitute $\rmd\bs{\psi}$ into Eq.~(\ref{eqn:Kin:diff_full_DT_Jacobian_defs}), arriving at the final differential kinematics:
\begin{equation}\label{eqn:Kin:diff_full_DT_Jacobians_final}
\rmd \bs{\xi} = \;\mb{J}_{M}\;\rmd\mb{q} \;+\; \mb{J}_\mu\;\rmd q_s \;+\; \mb{J}_\mb{k} \;\rmd \mb{k}_\lambda
\end{equation}
Equation (\ref{eqn:Kin:diff_full_DT_Jacobians_final}) fully decouples the end-effector pose differential to contributions of the direct (macro) actuation $\rmd\mb{q}$, the indirect (micro) actuation $\rmd q_s$, and the modeling uncertainty $\rmd\mb{k}_\lambda$. The three Jacobian matrices are obtained from Eq.~(\ref{eqn:Kin:diff_full_DT_Jacobians_final}) as an important finding of this paper: $\mb{J}_M$ defined as the \textit{Macro} motion Jacobian, $\mb{J}_\mu$ defined as the \textit{Micro} motion Jacobian, and $\mb{J}_\mb{k}$ defined as the \textit{Identification} Jacobian.
\begin{align}
	&\mb{J}_M = \left[\begin{array}{c;{2pt/2pt}c}
		\mb{J}_{\bs{\xi}\bs{\phi}} \,\mb{J}_{\bs{\phi}\theta}&\mb{J}_{\bs{\xi}\bs{\phi}} \,\mb{J}_{\bs{\phi}\delta} + \mb{J}_{\bs{\xi}\delta}
	\end{array}\right] (\mb{J}_{\mb{q}\bs{\psi}})^{\dagger},
	\;\mb{J}_M\in\realfield{6\times2} \label{eqn:Kin:diff_full_DT_Jacobian_M}\\
	&\mb{J}_\mu = \mb{J}_{\bs{\xi}\bs{\phi}} \,\mb{J}_{\bs{\phi}q_s}+\mb{J}_{\bs{\xi}q_s} ,
	\qquad \mb{J}_\mu\in\realfield{6\times1} \label{eqn:Kin:diff_full_DT_Jacobian_mu}\\
	&\mb{J}_\mb{k} = \mb{J}_{\bs{\xi}\bs{\phi}} \,\mb{J}_{\bs{\phi}\mb{k}_\lambda},
	\qquad  \mb{J}_\mb{k} \in\realfield{6\times n_k} \label{eqn:Kin:diff_full_DT_Jacobian_k}
\end{align}
Where $(\cdot)^{\dagger}$ is the Moore-Penrose pseudo inverse.

\subsection{Deriving $\mb{J}_{{\xi}{\phi}}$ and $\mb{J}_{{\xi}\delta}$}\label{ch:kin:2seg}
\par The result in Eq. (\ref{eqn:Kin:diff_full_DT_Jacobians_final}) builds on knowing the Jacobians $\mb{J}_{\bs{\xi}\bs{\phi}}\in\realfield{6\times2}$ and $\mb{J}_{\bs{\xi}\delta}\in\realfield{6\times1}$ as mentioned in Eqs. (\ref{eqn:Kin:Diff_full},~\ref{eqn:Kin:Diff_full_Jac}). We now provide a derivation to these two Jacobians. Considering a single-segment CREM as two concatenated subsegments (inserted and empty), we apply the Jacobian formulation for a two-segment multi-backbone continuum robot (MBCR) while assuming that both subsegments share the bending plane angle $\delta$. For the ease of adapting formulations from \cite{Xu2008}, we introduce a vector notation:
\begin{equation}\label{eqn:kin:delta_2_seg}
\bs{\delta}_v\triangleq
\begin{bmatrix}
\delta_s\\
\delta_\varepsilon
\end{bmatrix} =
\begin{bmatrix}
1 \\
1
\end{bmatrix}\delta, \quad
\rmd \bs{\delta}_v\triangleq
\begin{bmatrix}
\rmd \delta_s\\
\rmd \delta_\varepsilon
\end{bmatrix} =
\begin{bmatrix}
1 \\
1
\end{bmatrix}\rmd \delta
\end{equation}
\par\noindent We next use the notation of ${}^{i-1}\bs{\xi}_{i/i-1}$ to denote the pose of the $i^\text{th}$ subsegment relative to the $(i-1)^\text{th}$ subsegment where $i\in\{s, \varepsilon\}$. Using $\mb{v}$ and $\bs{\omega}$ to denote linear and angular velocities, we define the corresponding four Jacobian matrices corresponding with the contributions of $\rmd\theta_i, \rmd\delta_i$ where $i\in\{s, \varepsilon\}$ to the end-effector twist:
\begin{equation}\label{eqn:Jacobian_partitions}
\dfrac{\partial\; {}^{i-1}\bs{\xi}_{i/i-1}}{\partial \left([\theta_i,\delta_i]\T\right)}
\triangleq
\left[\begin{array}{c;{2pt/2pt}c}
\mb{J}_{\mb{v}\theta_i}
&\mb{J}_{\mb{v}\delta_i}
\vspace{1mm}\\\hdashline[2pt/2pt]
\mb{J}_{\bs{\omega}\theta_i}
&
\mb{J}_{\bs{\omega}\delta_i}
\end{array}\right]\in\realfield{6\times2}, \quad i\in\{s, \varepsilon\}
\end{equation}
Details of the derivations of the Jacobians are provided in Appendix \ref{app:jacobian_partitions}.
\par Following \cite{Xu2010}, the serial composition of two subsegments using twist transformation results in the end effector twist:
\begin{align}
	&
	\mb{J}_{\bs{\xi}\bs{\phi}} =
	\dfrac{\partial \bs{\xi}}{\partial \bs{\phi}} =
	\left[\begin{array}{c;{2pt/2pt}c}
		\mb{J}_{\mb{v}\theta_s} - \left[{}^b\mb{R}_c \,{}^{c}\mb{p}_{g/c}\right]^\wedge \mb{J}_{\bs{\omega}\theta_s} &
		{}^b\mb{R}_c \mb{J}_{\mb{v}\theta_{\varepsilon}} \\[2pt]\hdashline[2pt/2pt]
		\mb{J}_{\bs{\omega}\theta_s} & {}^b\mb{R}_c \mb{J}_{\bs{\omega}\theta_{\varepsilon}}
	\end{array}\right]\label{eqn:Kin:2seg:J_x_phi}\\
	&
	\mb{J}_{\bs{\xi}\bs{\delta}_v} =
	\dfrac{\partial \bs{\xi}}{\partial \bs{\delta}_v} =
	\left[\begin{array}{c;{2pt/2pt}c}
		\mb{J}_{\mb{v}\delta_s} - \left[{}^b\mb{R}_c \,{}^{c}\mb{p}_{g/c}\right]^\wedge \mb{J}_{\bs{\omega}\delta_s}&
		{}^b\mb{R}_c \mb{J}_{\mb{v}\delta_\varepsilon}\vspace{1mm}\\\hdashline[2pt/2pt]
		\mb{J}_{\bs{\omega}\delta_s} & {}^b\mb{R}_c \mb{J}_{\bs{\omega}\delta_\varepsilon}
	\end{array}\right]\label{eqn:Kin:2seg:J_x_delta_vec}
\end{align}
\par \noindent These definitions of $\mb{J}_{\bs{\xi}\bs{\phi}}$  and $ \mb{J}_{\bs{\xi}\bs{\delta}_v}$ complete the two missing terms needed in Eq. (\ref{eqn:Kin:diff_full_DT_Jacobians_final}), but with a slight formulation modification. The Jacobian matrix $\mb{J}_{\bs{\xi}\delta}$ is slightly different from $\mb{J}_{\bs{\xi}\bs{\delta}_v}$ in Eq.~(\ref{eqn:Kin:2seg:J_x_delta_vec}), and using the differentiation chain rule it becomes:
\begin{equation}\label{dd}
\mb{J}_{\bs{\xi}\delta}\triangleq
\dfrac{\partial \bs{\xi}}{\partial \delta}
\;
= \;  \dfrac{\partial \bs{\xi}}{\partial \bs{\delta}_v}\dfrac{\rmd\bs{\delta}_v}{\rmd\delta}
\;
= \;\mb{J}_{\bs{\xi}\bs{\delta}_v}
\begin{bmatrix}
1 \\
1
\end{bmatrix}
\end{equation}

\section{Calibration of Micro Motion Parameters}\label{ch:calibration}
To calibrate the model uncertainty parameters $\mb{k}_\lambda$, we extract from Eq.~(\ref{eqn:Kin:diff_full_DT_Jacobians_final}) the following relation:
\begin{equation}\label{dd}
\delta \bs{\xi} (\delta\mb{k}_\lambda) = \mb{J}_\mb{k} \delta\mb{k}_\lambda
\end{equation}
Using this error propagation model, we construct an estimation method to estimate $\mb{k}_{\lambda}$.  Let $\bs{\xi}_j\leftrightarrow \left[\bar{\mb{x}}_{j},\bar{\mb{R}}_{j}\right]$ designate the measured end-effector pose at the $j^\text{th}$ robot configuration (insertion depth) where $\bar{\mb{x}}_{j}$ and $\bar{\mb{R}}_{j}$ designate the measured position and orientation. Let $\mb{x}_{j}$ and $\mb{R}_{j}$ denote the modeled pose using the direct kinematics as presented in section \ref{ch:kin:kin} for a given $\mb{k}_{\lambda}$. The error between the measured and modeled poses are then defined as:
\begin{equation}\label{eqn:cal:error_function_multi_ij}
\mb{c}_j \triangleq \left[(\bar{\mb{x}}_{j}-\mb{x}_{j})\T,(\alpha_{e_{j}}\uvec{m}_{e_{j}})\T\right]\T, \quad  \mb{c}_{j}\in\realfield6
\end{equation}
where $\alpha_{e_j}$ and $\uvec{m}_{e_j}$ are the angle and axis parameterizing the orientation error $\mb{R}_{e_j}$. These parameters are given by:
\begin{align}
	&\mb{R}_{e_j}\triangleq\bar{\mb{R}}_j{\mb{R}_j}\T=e^{\alpha_{e_j}\,[\uvec{m}_{e_j}]^{\wedge}}\label{eqn: Cal:diff_R}\\
	&  \alpha_{e_j} = \cos^{-1}\left(\tfrac{\text{Tr}(\mb{R}_{e_j})-1}{2}\right)
	\label{eqn:Cal:diff_R_1}\\
	& \uvec{m}_{e_j} = \dfrac{1}{2\sin(\alpha_e)}
	\left(\mb{R}_{e_j}-{\mb{R}_{e_j}}\T\right)^\vee
	\label{eqn:Cal:diff_R_2}
\end{align}
where the operator $(\cdot)^\vee$ designates the vector form of a skew-symmetric matrix.
\par An \textit{aggregated} error vector $\widetilde{\mb{c}}_{\lambda}$ is defined to include errors of all $N$ robot configurations:
\begin{equation}\label{eqn:Cal:aggregate_error_vec}
\widetilde{\mb{c}}_\lambda=\left[(\mb{c}_{1})\T,\hdots,(\mb{c}_{N})\T\right]\T
\end{equation}
The optimization objective function $M_\lambda$ is then defined as:
\begin{equation}\label{eqn:Cal:M_gamma}
M_\lambda(\mb{k}_\lambda) = \frac{1}{2N}\;\widetilde{\mb{c}}_\lambda{}\T \;\mb{W} \;\widetilde{\mb{c}}_\lambda
\end{equation}
Where $\mb{W}$ is a weight matrix encoding confidence in the measurements and the measurement unit scaling factors.
\par The first-order Taylor series approximation of $M_\lambda$ is given:
\begin{equation}\label{eqn:M_lambda_taylor}
M_\lambda(\mb{k}_\kappa+\delta\mb{k}_\kappa)\approx M_\lambda(\mb{k}_\lambda)+\mb{J}_{M_\lambda}\delta\mb{k}_\lambda
\end{equation}
where \textit{the aggregated} Jacobian $\mb{J}_{M_\lambda}\in \realfield{1\times5}$ is given by:
\begin{align}
	&\mb{J}_{M_\lambda}=\frac{1}{N}(\widetilde{\mb{c}}_\lambda)\T\;\mb{W}\;\mb{J}_{c_\lambda}
	\label{eqn:cal_1_obj_func_Jacobian_M_lambda}\\
	&\mb{J}_{c_\lambda} =
	\frac{\partial \widetilde{\mb{c}}_\lambda}{\partial \mb{k}_\lambda} =
	-
	\left [ \left(\mb{J}_{\mb{k}_1}\right)\T,
	\hdots,
	\left(\mb{J}_{\mb{k}_N}\right)\T
	\right]\T\label{eqn:cal_1:pC_pk_curv}
\end{align}
\par Equation~(\ref{eqn:cal_1_obj_func_Jacobian_M_lambda}) shows that minimizing $M_\lambda$ entails following the gradient descent direction along $({\partial \widetilde{\mb{c}}_\lambda}/{\partial \mb{k}_\lambda})$. The parameter $\mb{k}_\lambda$ is then obtained using an iterative nonlinear least squares estimation shown in Algorithm \ref{algo:crem:NLS}.
\begin{algorithm}
	\caption{Nonlinear LS Estimate}
	\label{algo:crem:NLS}
	\begin{algorithmic}[1]
		\footnotesize
		\Require $\mathcal{D}\{(\bar{\mb{x}}_j,\bs{\psi}_j,q_{s_j})\}$, $_{j=1,\hdots,N}$;
		$\mb{k}_{\lambda_0}$, $(\beta,\eta)>0$
		\State \textbf{START} Initialize: $\mb{k}_i \gets \mb{k}_{\lambda_0}$,\hspace{2mm}$M_{i-1} \gets 1$,\hspace{2mm} $M_i \gets 100$
		\While{$\frac{\|{M_\lambda}_i - {M_\lambda}_{i-1}\|}{{M_\lambda}_{i-1}}\geq \beta$}
		\State ${M_\lambda}_{i-1} \gets {M_\lambda}_i$, \hspace{2mm}
		$\widetilde{\mb{c}}_p= \widetilde{\mb{c}}_\lambda(\mb{k}_i)$,\hspace{2mm}
		${M_\lambda}_i = {M_\lambda}_i(\mb{k}_i)$,\hspace{2mm}
		\State $\mb{J}_{c_\lambda} = \mb{J}_{c_\lambda}(\mb{k}_i)$
		\State Update $\mb{k}_{i+1}$:
		\begin{align}
			&\mb{k}_{i+1} = \mb{k}_i - \mb{H}\;\left(\eta \,\left(\mb{J}_{c_\lambda}\right)^{\boldsymbol{+}}\,\widetilde{\mb{c}}_\lambda\right), \eta\in(0, 1]
			\label{eqn:algo:NLS_update_k}\\
			&\left(\mb{J}_{c_\lambda}\right)^{\boldsymbol{+}} =
			\left((\mb{J}_{c_\lambda})\T \,\mb{W} \,\mb{J}_{c_\lambda}\right)^{-1}
			(\mb{J}_{c_\lambda})\T \,\mb{W}
			\label{eqn:algo:J_M_plus}
		\end{align}
		\EndWhile
		\State $\mb{k}^{*} \gets \mb{k}_i$
		\Ensure ${\mb{k}^{*}}$
	\end{algorithmic}
\end{algorithm}
%
\par In the algorithm, $\mb{H}$ is the parameter scaling matrix and the task space variable scaling is achieved by adjusting $\mb{W}$, both of witch are discussed in details in \cite{siciliano2008springer}. 

\section{Simulation Study of Direct Kinematics and Differential Kinematics}
\par In this section, we present simulations to demonstrate the direct kinematics and differential kinematics. We also verify the differential kinematics through finite-difference simulations. We also carry out simulations to verify the differential kinematics model. In all simulations, we assumed the robot points vertically down at its home (straight) configuration. 
\subsection{Position Analysis of Micro Motion}\label{ch:simu:lambda0}
\begin{figure*}[htb]
	\centering
	\includegraphics[width= 0.95\linewidth]{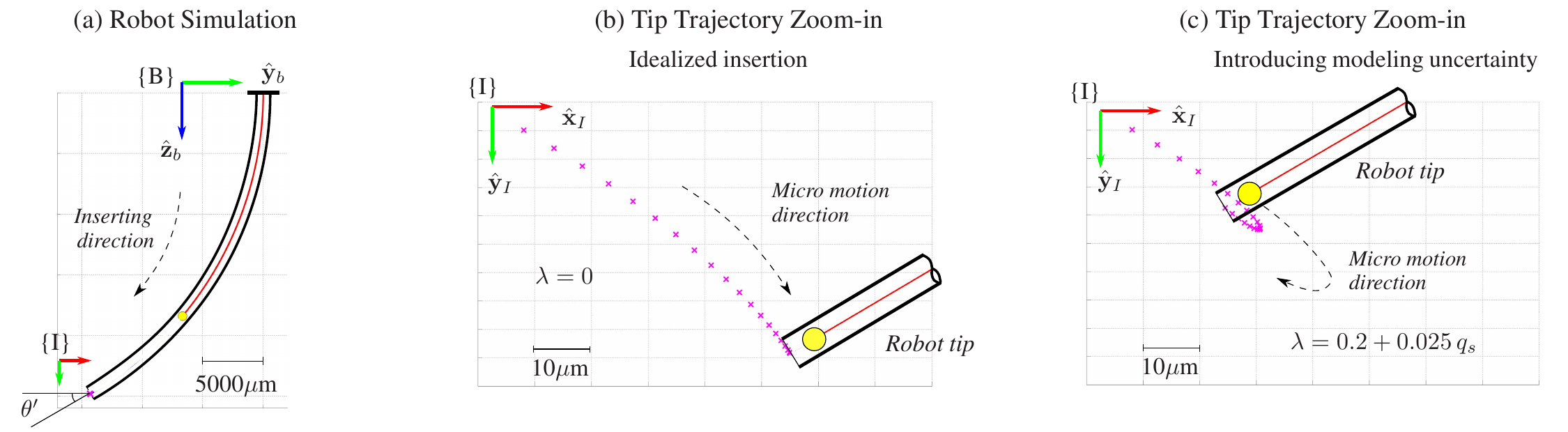}
	\caption{Simulations of continuum robot micro motion created by EMB insertion: (a) the entire segment when EMB being inserted (only central backbone and EMB are shown); (b) zoom-in view of the tip micro motion assuming ideal condition, i.e. $\lambda=0$; (c) zoom-in view of the tip micro motion assuming a linear uncertainty function of $\lambda$. $\{\text{B}\}$ and $\{\text{I}\}$ designate the robot base and the camera image frames.}
	\label{fig:micro_ins_robot_tip}
\end{figure*}
\par Using the model in section \ref{ch:kin:kin}, we present the simulated position analysis of the micro motion created by the EMB insertion. In both simulations and the experimental model validations, we use the parameters as in Table~\ref{tab:simulation_param}. They include the Young's modulus of the superelastic NiTi material used for backbone tubes and EMBs ($E_p$, $E_i$, $E_s$), the diameters of backbones ($d_p$, $d_i$, $d_s$), and the cross-sectional moment of inertia ($I_p$, $I_i$, $I_s$).
%
\begin{table}[htbp]
	\centering
	\caption{Robot Parameters Used in Simulations and Experiments}
	\label{tab:simulation_param}
	\footnotesize
	\vspace{1mm}
	\setlength\tabcolsep{2.5pt}
	\begin{tabular}{@{}ccccccc@{}}  
		L & r & $E_p , E_i, E_s$ & $d_p , d_i $ & $d_s$       & $I_p, I_i$                          & $I_s$  \\ \thickhline{1.5pt}
		44.3mm & 3mm & 41 GPa & 0.90 mm  & 0.38 mm & 0.0312 $\text{mm}^4$  & 0.0010  $\text{mm}^4$\\ \thickhline{1pt}
	\end{tabular}
\end{table} 
%
\par \noindent Figure~\ref{fig:micro_ins_robot_tip} shows the simulation results of the micro motion created by EMB insertion. Figure~\ref{fig:micro_ins_robot_tip}(a) shows the continuum segment at its initial bending angle $\theta=30^\circ$ . During simulation, the equilibrium bending angles $\{\theta', \theta_s\}$ were computed at different EMB insertion depths. The resulting tip micro-motion is shown in Fig.~\ref{fig:micro_ins_robot_tip}(b) for the na\"ive kinematic model (i.e. $\lambda=0$). Figure ~\ref{fig:micro_ins_robot_tip}(c) shows the tip motion for an updated model assuming \cusst{$\lambda=0.2+0.06q_s$}\corrlab{R2-4}{$\lambda=0.2+0.025q_s$}. We note that, as expected, in both cases the robot straightens with EMB insertion since the robot body straightens. However, the updated model exhibits a turning point behavior which relates to the combined effect of straightening and change in the end effector angle $\theta'$. This same phenomenon was observed experimentally in section \ref{ch:exp}. \corrlab{R2-2}{The particular values used for the parameters defined in Eq.~(\ref{eqn:model_uncertainty_def}), $k_{\lambda_0}=0.2, k_{\lambda_q}=0.025$, are manually selected to illustrate the turning point behavior that is similar to what is observed in experiments. In practise, they are calibrated according to section \ref{ch:exp:calibration}.}
\subsection{Instantaneous Kinematics and Error Propagation}\label{ch:simu:Jacobian}
\par To verify the derivations of instantaneous kinematics and error propagation, we compute Jacobians following section \ref{ch:kin:differential}. Since the simulation case represents the robot motion within a bending plane, the columns of the Jacobians represent $2\times1$ vectors of induced velocities for unit change in the variables associated with each Jacobian. The following simulations verify the macro motion Jacobian $\mb{J}_M$, the micro motion Jacobian $\mb{J}_\mu$, and the identification Jacobian $\mb{J}_\mb{k}$ by plotting the Jacobian columns. The verification is carried out visually by verifying that the Jacobian columns induce tip velocity tangent to the trajectory generated by direct kinematics.  In addition, each Jacobian has been also verified numerically via finite difference computations.
\par
\begin{figure}[htb]
	\centering
	\includegraphics[width= 0.99\columnwidth]{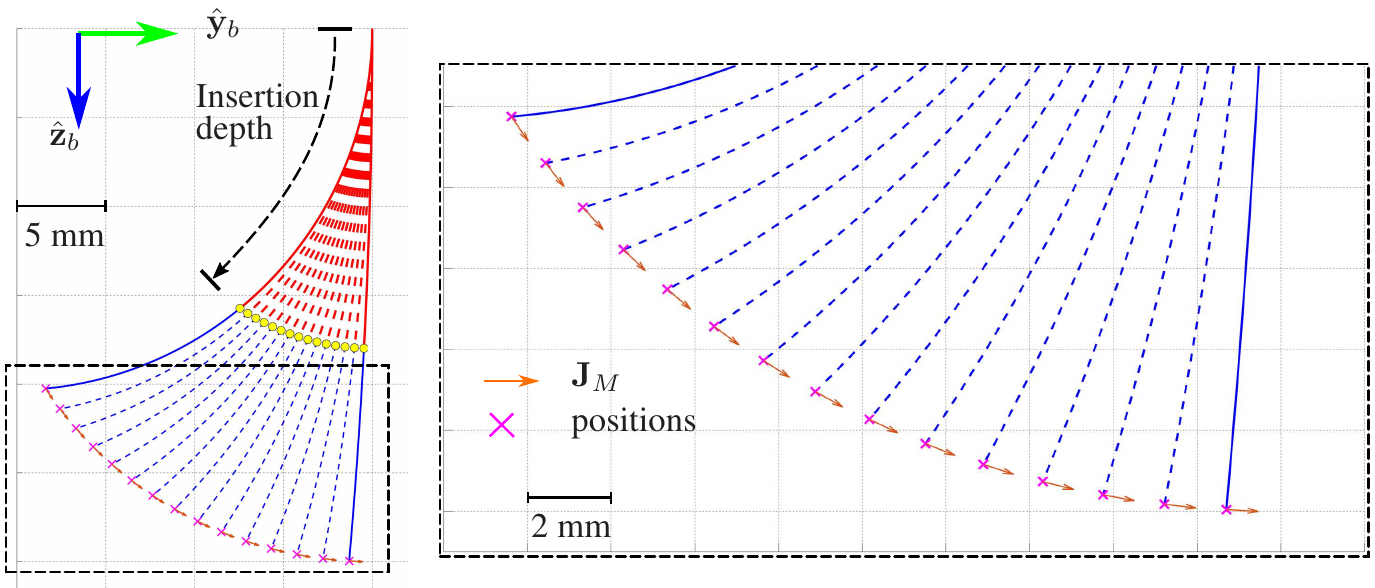}
	\caption{The macro motion simulation and the micro motion Jacobian computed during the simulation (red indicates inserted portion, blue indicates empty portion). The Jacobian $\mb{J}_M$ is shown in brown arrows representing induced tip velocities.}
	\label{fig:simu_jacobians_macro}
\end{figure}

 To verify $\mb{J}_M$, the EMB insertion depth $q_s$ was fixed and direct actuation of backbones was assumed. Sample tip positions along the trajectory were obtained via direct kinematics and the corresponding Jacobian $\mb{J}_M$ was computed. Figure~\ref{fig:simu_jacobians_macro} shows the simulation results. These results verify that the computed $\mb{J}_M$ is tangent to the macro motion trajectory.
\par
\begin{figure*}[htb]
	\centering
	\includegraphics[width=  0.99\linewidth]{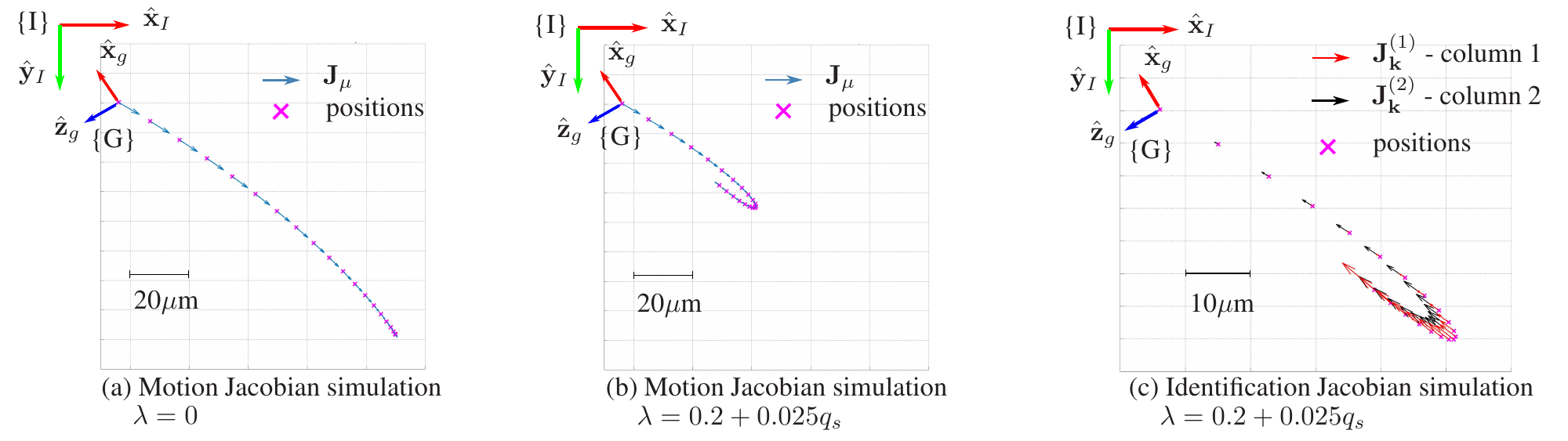}
	\caption{Simulations verifying derivations of Jacobians: (a) micro motion Jacobian when $\lambda=0$; (b) micro motion Jacobian when $\lambda\neq0$; (c) error propagation using $\mb{J}_\mb{k}$ and $\rmd \mb{k}_\lambda$ where perturbations were overlapped in simulated trajectory.}
	\label{fig:simu_jacobians_micro_J_k}
\end{figure*}

 To verify $\mb{J}_\mu$, the secondary backbones were assumed locked and the EMB insertion depth $q_s$ was varied. The Jacobian $\mb{J}_\mu$ was computed and plotted for each EMB depth. Two different scenarios of modeling uncertainty were considered: $\lambda=0$ and \cusst{$\lambda=0.2+0.06q_s$}\corrlab{R2-4}{$\lambda=0.2+0.025q_s$}.  The results in Fig.~\ref{fig:simu_jacobians_micro_J_k}(a)(b) verify that $\mb{J}_\mu$ is tangent to the micro scale trajectory generated by direct kinematics.
\par Figure~\ref{fig:simu_jacobians_micro_J_k}(c) shows the plots of the identification Jacobian $\mb{J}_\mb{k}$ for the simulation scenario where $\lambda\neq0$, revealing how the parameter errors of modeling uncertainty affect the tip positions and hence the shape of the trajectory.

\section{Experimental Validations}\label{ch:exp}
In \cite{delgiudice2017IROS_submitted} the feasibility of micro motion through equilibrium modulation was demonstrated. The following experiments evaluate the ability of our simplified kinematic model to capture the micro-motion behavior, validate the calibration framework in section \ref{ch:calibration}, and assess the accuracy of the updated kinematic model in reflecting the experimental data.
\subsection{Experimental Setup \& Ground Truth Data}\label{ch:exp:setup}
\begin{figure}[htbp]
	\centering
	\includegraphics[width= 0.98 \columnwidth]{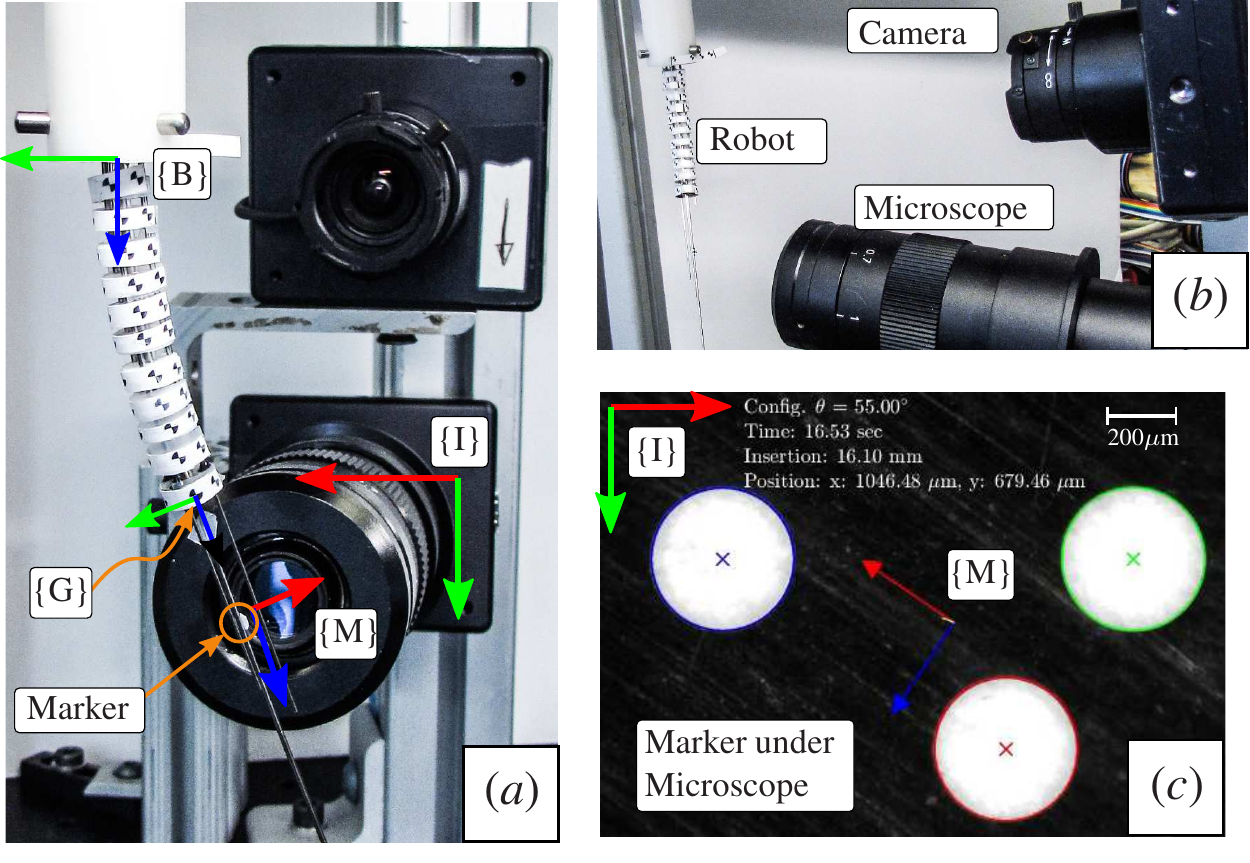}
	\caption{Experimental Setup: (a) a single-segment continuum robot whose motion is captured by two cameras; (b) the side view of the setup; (c) the segmented marker under the microscope view.}
	\label{fig:exp_setup}
\end{figure}
A single-segment continuum robot with EMB insertion actuation was used as the experimental platform, Figure~\ref{fig:exp_setup}. The platform was presented in \cite{delgiudice2017IROS_submitted}, and it was modified from an earlier multi-backbone continuum robot design \cite{Xu2008}. To observe the robot tip motion at different scales, one HD camera (FLIR Dragonfly II\textsuperscript\textregistered) was used to capture the macro motion and the bending shape while an identical camera mounted on a $22.5\times$ microscope lens to capture the micro motion. Custom ``multi-circled'' marker was used to track the tip motion under microscope while multiple custom ``X'' markers were attached to the continuum robot body to observe the bending shape. The vision measurement methods used were presented in \cite{delgiudice2017IROS_submitted} with the micro motion tracking accuracy being reported better than 2 $\mu$m.
\par Fig.~\ref{fig:exp_setup} shows the frames used and also previously referred to in Fig.~\ref{fig:micro_ins_robot_tip}. The microscope is fixed at a known offset relative to the robot base, and such offset is represented as a constant transformation from the image frame \{I\} to the robot base frame \{B\}. The tracked marker frame \{M\} is placed at a known offset relative to the end disk (gripper frame \{G\}), and the transformation is represented as a constant transformation between  \{M\} and \{G\}. The marker position and orientation in the image frame is obtained by the segmentation of the three circles that construct an asymmetric pattern, as shown in  Fig.~\ref{fig:exp_setup}(c).
\begin{figure}[htbp]
	\centering
	\includegraphics[width= 0.75 \columnwidth]{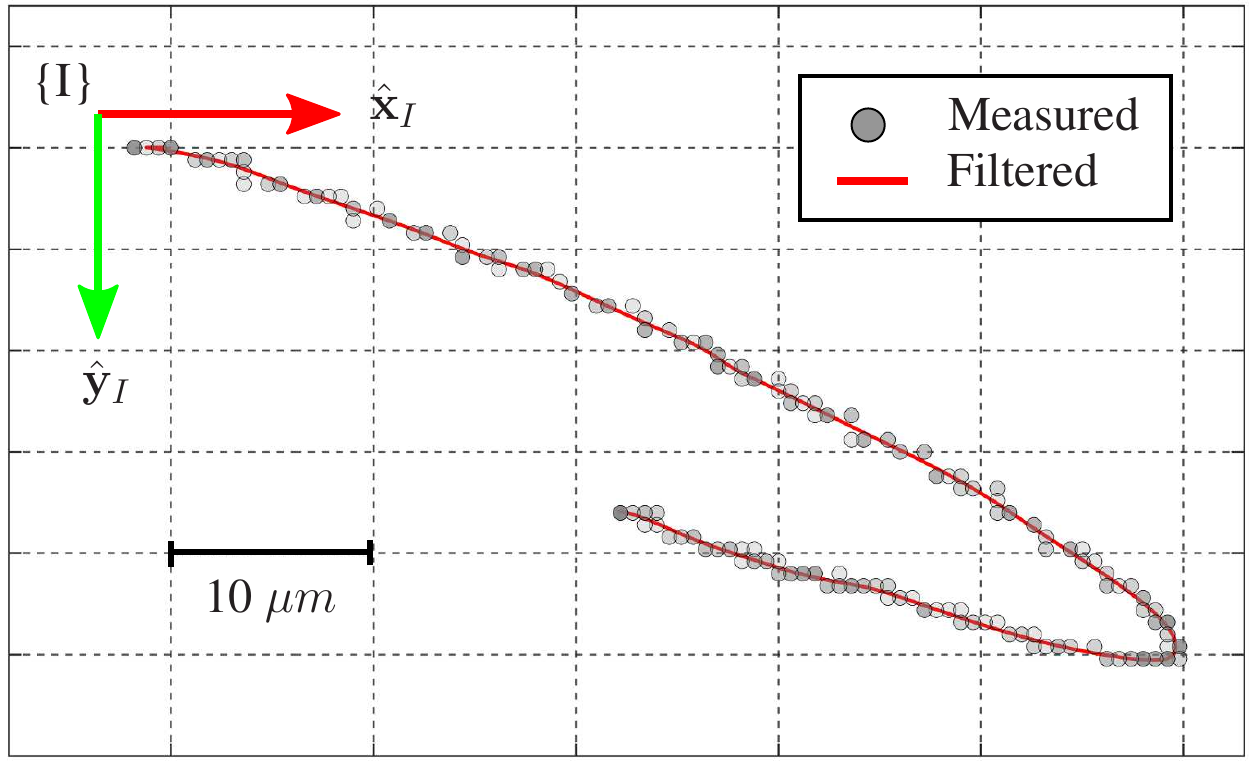}
	\caption{One example of image-segmented tip positions under microscope when macro bending angle $\theta=45^\circ$}
	\label{fig:micro_insertion_experiment}
\end{figure}
\par
\begin{figure}[htbp]
	\centering
	\includegraphics[width= 0.75 \columnwidth]{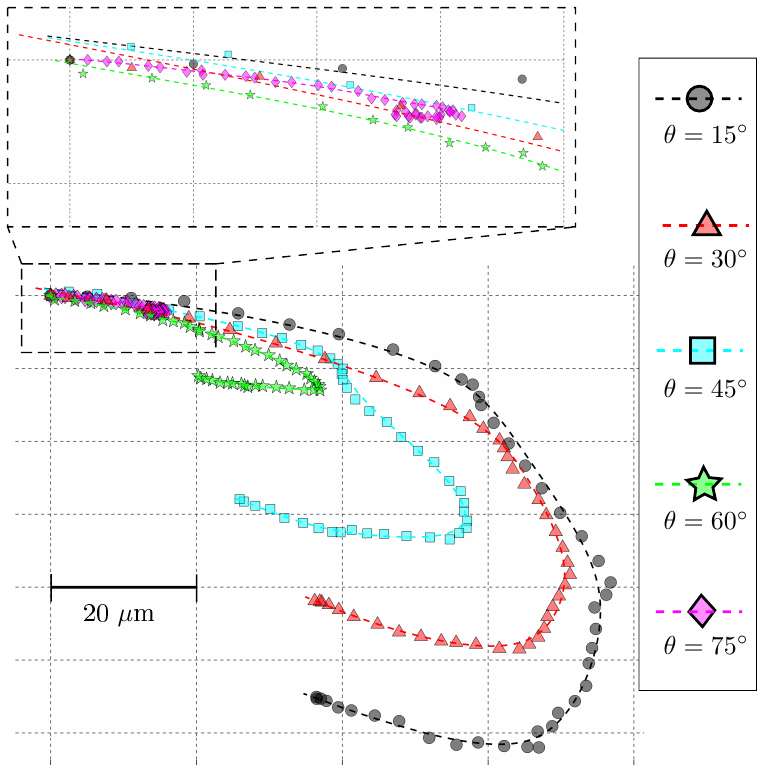}
	\caption{\corrlab{R1-2C}{Averaged micro motion trajectories under microscope when the macro bending angles $\theta=15^\circ, 30^\circ, 45^\circ, 60^\circ, 75^\circ$. Each plotted trajectory is obtained by averaging over 10 repetitions of the EMB insertion experiment at the particular macro bending angle.}}
	\label{fig:micro_insertion_experiment_overlay}
\end{figure}

 The \href{http://arma.vuse.vanderbilt.edu/images/stories/videos/LongJMR2018.mp4}{multimedia extension} and Fig.~\ref{fig:micro_insertion_experiment} show a sample marker frame trajectory during EMB insertion \corrlab{R1-2A}{where the macro motion bending angle was chosen to be $\theta=45^\circ$}. The marker positions were segmented from microscope images collected at 30 frames per second. Applying a butterworth infinite impulse response filter with the 3-dB frequency as 30 Hz, provided a smooth trajectory for calibration. \corrlab{R1-2B}{Results of more experiments are plotted in Fig.~\ref{fig:micro_insertion_experiment_overlay}: five macro bending angle configurations, $\theta=15^\circ, 30^\circ, 45^\circ, 60^\circ, 75^\circ$, were chosen to sample the workspace, and for each value of $\theta$, ten repetitions of the EMB insertion experiment were conducted. Figure~\ref{fig:micro_insertion_experiment_overlay} shows that the turning point phenomenon exists over the entire workspace.}

\subsection{Model Calibration}\label{ch:exp:calibration}
\par Using calibration method in section \ref{ch:calibration}, we calibrated the modeling uncertainty parameter vector $\mb{k}_\lambda$ . The parameter vector $\mb{k}_\lambda$ in Eq.~(\ref{eqn:model_uncertainty_def}) consists of three elements, a bias term $k_{\lambda_0}$, a coefficient gain $k_{\lambda_\theta}$ that is associated with the nominal bending angle $\theta$, and a coefficient gain $k_{\lambda_q}$ that relates to the EMB insertion depth $q_s$. As a preliminary study, in this paper, we focus on investigating and calibrating $k_{\lambda_0}$ and $k_{\lambda_q}$. \corrlab{R2-3}{To obtain the value of $k_{\lambda_\theta}$, there are multiple possible solutions. In one method, after collecting observation data of micro motion trajectories from sufficient groups varying bending configuration $\theta$, one could use the same calibration method with the identification Jacobian $\mb{J}_\mb{k}$ that includes the third column corresponding to $k_{\lambda_\theta}$. In another method, one could first calibrate $k_{\lambda_0}$ and $k_{\lambda_q}$ for each micro motion trajectory that is observed in a particular bending configuration $\theta$, followed by generating an interpolating lookup table in the format of $(k_{\lambda_0}, k_{\lambda_q})=f_\text{Lookup}(\theta)$.} \cusst{Once the characterization of $k_{\lambda_0}$ and $k_{\lambda_q}$ is achieved, one can exhaust the choices of  $\theta$ to investigate the effect of $k_{\lambda_\theta}$.}
\par Algorithm \ref{algo:crem:NLS} was initialized with $k_{\lambda_0}=0$, $k_{\lambda_q}=0$. In each iteration, the modeled positions were computed using the current estimates of the parameters. The \textit{aggregated} error vector was then calculated between the modeled and experimental positions. For each iteration, both of the current estimates of the parameters and the position root-mean squared errors (RMSE) of all insertion samples (382 in total) were reported. A relative convergence threshold of  $0.1\%$ was used to determine the convergence.
\par
\begin{figure*}[htbp]
	\centering
	\includegraphics[width= 0.95 \linewidth]{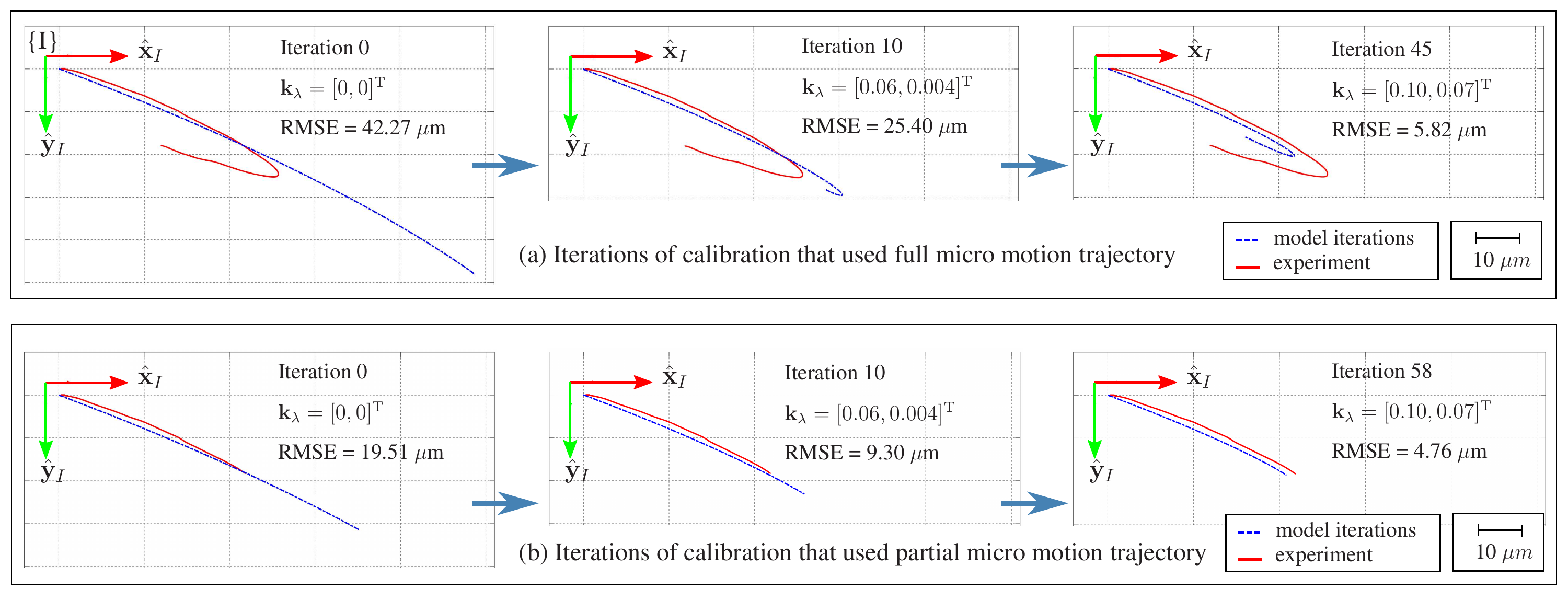}
	\caption{Experimental data and model iterations during the parameter estimation (model calibration).}
	\label{fig:micro_insertion_calibration}
\end{figure*}

 For the particular experimental data collection shown in Fig.~\ref{fig:micro_insertion_experiment}, the parameter estimation (model calibration) went through 46 iterations before converging, where a step size of $\eta=0.1$ was selected ($\eta$ as introduced in Algorithm \ref{algo:crem:NLS}). Figure~\ref{fig:micro_insertion_calibration}(a) shows selected iterations during the estimation, and the details of the iterations are reported in Table~\ref{tab:calibration_results}. The estimation started with an initial position RMSE of 44.27 $\mu$m, and after its convergence, the position RMSE was reduced to 5.82 $\mu$m, showing an improvement of 86.8\% in model errors.
\par By dividing the tip trajectory into two segments, we observe that the current simplistic modeling approach produced bigger errors after the turning point: the RMSEs were reported as $4.87 \mu m$ and $6.63 \mu m$ for the two segments before and after turning point that had the lengths of $48.11 \mu m$ and $38.82 \mu m$, respectively. If one wishes to further improve the model accuracy, a model that only predicts the trajectory before the turning point may be considered. We therefore considered another calibration where only the partial micro motion trajectory before the turning point was used. With the same iteration step size and convergence criterion, the estimation went through 59 iterations to converge, and the updated results were reported in Table~\ref{tab:calibration_results_turning_point} and plotted in Fig.~\ref{fig:micro_insertion_calibration}(b). The position RMSE was then further improved to $4.76 \mu m$.
%
\begin{table}[htbp]
	\centering
	\caption{Calibration using full micro motion trajectory}
	\label{tab:calibration_results}
	\footnotesize
	\vspace{1mm}
	\begin{tabular}{@{}clllllll@{}}
		Iteration                                                     & 0         & 5          & 10     & 20      & 30     & 45        & 46    \\ \thickhline{1.5pt}
		\rule{0pt}{1.3\normalbaselineskip}$\dfrac{k_{\lambda_0}}{100}$     & 0         & 4.22    &6.72  & 9.06  & 9.88  & 10.23  & 10.24 \\
		\rule{0pt}{1.6\normalbaselineskip}$\dfrac{k_{\lambda_q}}{1000}$     & 0         & 2.7      &4.3     & 5.7     & 6.3    & 6.5       & 6.5 \\
		RMSE [$\mu$m] & \textbf{42.27}      & 15.71 & 25.40 &7.72  & 6.07  & 5.82    & \textbf{5.82}  \\ \thickhline{1pt}
	\end{tabular}
\end{table} 
\vspace*{-2.0\baselineskip}
\begin{table}[htbp]
	\centering
	\caption{Calibration using partial micro motion trajectory}
	\label{tab:calibration_results_turning_point}
	\footnotesize
	\vspace{1mm}
	\begin{tabular}{@{}clllllll@{}}
		Iteration                                                     & 0         & 5          & 10      & 20      & 30     & 58        & 59    \\ \thickhline{1.5pt}
		\rule{0pt}{1.3\normalbaselineskip}$\dfrac{k_{\lambda_0}}{100}$     & 0         & 4.22    & 6.72   & 9.06  & 9.88  & 10.30  & 10.30 \\
		\rule{0pt}{1.6\normalbaselineskip}$\dfrac{k_{\lambda_q}}{1000}$     & 0         & 2.7      & 4.3      & 5.7     & 6.3    & 6.5       & 6.5 \\
		RMSE [$\mu$m] & \textbf{19.51}                      & 13.01 & 9.30 &6.12  & 5.17  & 4.76    & \textbf{4.76}  \\ \thickhline{1pt}
	\end{tabular}
\end{table} 
%
\vspace*{-1.0\baselineskip}
\subsection{{Limitations}}
\par This work focused on creating a simplistic, yet fast model for equilibrium modulation control implementation. \remind{O-1}{The kinematic model traded accurate mechanics modeling (which typically leads to solving nonlinear boundary value problems) with simplicity and speed of computation.} \corrlab{R1-3}{In this simplistic model, one EMB is assumed to be inserted through the central backbone, or equivalently, multiple EMBs are assumed to be inserted in coordination such that a shared insertion separation plane can be defined (shown as Fig.~\ref{fig:SnakeSpring}). Hence the current method limits the use of modeled micro motion to be coplanar with the macro bending plane, making the controllable micro motion have only one degree of freedom (DoF). A more sophisticated modeling method could potentially capture any arbitrary insertions of multiple EMBs, which enables the full capability of the micro motion mechanism that can generate spatial motions, for example, spiral motion.}
\par \remind{O-1}{Our experimental data showed an unexpected motion behavior manifested by a turning point along the micro-motion trajectory. The model presented in this paper does not offer a physical explanation to this behavior, but can capture this behavior for a given robot.} The model calibration results indicate that there is still a potential to improve the model performance by further investigating alternative modeling assumptions and different descriptions of modeling uncertainties. One of the limitations of our approach can be inferred from the simulation shown in Fig.~\ref{fig:simu_jacobians_micro_J_k}, where both columns of the identification Jacobian are almost aligned with the tangent to the direct kinematics trajectory. The attainable correction directions that the column-space of $\mb{J}_\mb{k}$ affords is therefore limited in reshaping the model trajectory. This was also observed from the experimental validation. Shown from the iterations in Fig.~\ref{fig:micro_insertion_calibration}, it is difficult to reshape the modeled tip trajectory in the direction that is perpendicular to the trajectory. The other limitation is potentially caused by the choice of linearity in expressing the modeling uncertainty, which may not be descriptive enough. 

\vspace{-1.0\baselineskip}
\section{Conclusion}
\par  This work presented the first modeling attempt for a new class of continuum robots capable of multi-scale motion. These robots achieve macro-scale and micro-scale motions through direct and indirect actuation (equilibrium modulation). Instead of focusing on a high-fidelity mechanics-based model, which typically leads to non-linear boundary value problems not easy to adopt for real-time control or parameter identification, this paper presented a simplified mechanics-based model utilizing moment coupling effects between sub-segments of the continuum robot. This approach generates a differential kinematics model that covers both macro and micro-motion. As a result of unavoidable parameter uncertainty, we presented a model-calibration approach that can compensate for parameter inaccuracy, friction effects and modeling inaccuracies due to the simplistic modeling assumptions. The modeling approach along with the calibration framework was validated experimentally on a multi-backbone continuum robot. The calibrated model reported a positional root-mean-squared error as 5.83 $\mu$m if one wishes to use the model for the entire motion profile with the turning point. If one chooses to exclude motions past the turning point, the calibrated model fit the experimental data with an accuracy of 4.76 $\mu m$. Future work will focus on investigations on a more sophisticated models capable of incorporating geometric constraints as well as minimizing mechanical energy for improved model accuracy. In addition, effects of direction reversal of EMB insertion can manifest in hysteresis, which has not been explored in this work, but still remains the topic of ongoing research. 
\section{Multimedia Extension}
The video shows the micro-motion trajectory during EMB insertion. The following is a link that would ideally be placed and linked on AMSE website if the paper is accepted. If ASME has no way of hosting these videos, then we can maintain them on our website.
\par \noindent \textbf{Multimedia}
\href{https://youtu.be/4GlLQwXUJpI}{https://youtu.be/4GlLQwXUJpI}.
\begin{acknowledgment}
This work was supported by NSF grant \#CMMI-1537659. Any opinions, findings, and conclusions
or recommendations expressed in this material are those of the author(s) and do not necessarily reflect the views of the National Science Foundation.
\end{acknowledgment}
%
\bibliographystyle{asmems4}
\bibliography{bib/crem_bib}%
\balance
\appendix
\numberwithin{equation}{section}
\makeatletter
\newcommand{\section@cntformat}{Appendix \thesection:\ }
\makeatother

\appendix
\section{Deriving $\tfrac{\partial {\phi}}{\partial \theta}$, $\tfrac{\partial {\phi}}{\partial \delta}$, $\tfrac{\partial {\phi}}{\partial q_s}$ \& $\tfrac{\partial {\phi}}{\partial \mb{k}_\lambda}$}\label{app:d_phi}
\par Rewriting Eq.~(\ref{eqn:SnakeSpring_theta_s_uncertainty}) and Eq.~(\ref{eqn:SnakeSpring_theta_prime}) in a matrix form yields:
\begin{equation}\label{eqn:App:theta_s_theta_prime}
\underbrace{\left[\begin{array}{cc}
	k_1+k_2+k_s & -k_1 \\
	k_1 & -k_1
	\end{array}
	\right]}_{\triangleq\;\mb{A}(\bs{\psi},q_s,\theta_s)} \;
\left[\begin{array}{c}
\theta_s \\
\theta'
\end{array}
\right]\;
= \;
\underbrace{\left[\begin{array}{c}
	(k_2 + k_s)\,\theta_0 - \lambda \\
	k_0(\theta_0-\theta)
	\end{array}
	\right]}_{\triangleq\;\mb{B}(\bs{\psi},q_s,\mb{k}_\lambda,\theta_s)}
\end{equation}
where $\mb{A}$ and $\mb{B}$ are defined as functions of $\{\bs{\psi}, q_s, \theta_s\}$ and $\{\bs{\psi}, q_s, \mb{k}_\lambda, \theta_s\}$ respectively. Using the definition of $\bs{\phi}$, yields:
\begin{equation}\label{eqn:App:phi_temp}
\bs{A}
\left(\underbrace{\left[\begin{array}{cc}
	1 & 0 \\
	1 & 1
	\end{array}
	\right]}_{\triangleq\;\mb{S}_0}
\underbrace{\left[ \begin{array}{c}
	\theta_s \\
	\theta_{\varepsilon}
	\end{array}
	\right]}_{\bs{\phi}} -
\underbrace{\left[ \begin{array}{c}
	0 \\
	\theta_0
	\end{array}
	\right]}_{\triangleq\; \mb{C}_0}
\right) = \mb{B}
\end{equation}
By introducing two constant matrices in the above equation, $\mb{S}_0$ and $\mb{C}_0$, we have obtained the equation to differentiate:
\begin{equation}\label{eqn:App:phi}
\mb{A}\,\left(\mb{S}_0\;\bs{\phi}  - \mb{C}_0\right)= \mb{B}, \quad \mb{A}\in\realfield{2\times2}, \mb{B}\in\realfield{2}, \mb{C}_0\in\realfield{2}
\end{equation}
The full differentiation may be expressed as:
\begin{equation}\label{eqn:App:d_eqn_phi}
(\rmd \mb{A}) \;\left(\mb{S}_0\,\bs{\phi} - \mb{C}_0\right)+ (\mb{A}\mb{S}_0) \;\rmd \bs{\phi} = \rmd \mb{B}
\end{equation}
Using $\mb{X}'_a$ to denote the partial derivative of matrix $\mb{X}$ w.r.t the scalar variable $a$, i.e. $\mb{X}'_a\triangleq\tfrac{\partial \mb{X}}{\partial a}$, then $\rmd \mb{A}$ and $\rmd \mb{B}$ may be written as:
\begin{align}
	& \rmd  \mb{A} =
	\mb{A}'_\theta \,\rmd \theta  +
	\mb{A}'_\delta \,\rmd \delta  +
	\mb{A}'_{q_s} \,\rmd q_s  +
	\mb{A}'_{\theta_s} \,\rmd \theta_s \label{eqn:App:dA}\\
	& \rmd \mb{B} =
	\mb{B}'_\theta \,\rmd \theta  +
	\mb{B}'_\delta \,\rmd \delta  +
	\mb{B}'_{q_s} \,\rmd q_s  +
	\mb{B}'_{\theta_s} \,\rmd \theta_s  +
	\sum\nolimits_{i}^{n_k}\mb{B}'_{k_{\lambda_i}}\rmd k_{\lambda_i} \label{eqn:App:dB}
\end{align}
Let us define $\mb{C}_{\bs{\phi}}$ and $\bs{\Gamma}_a$ to provide ease in the derivations:
\begin{equation}\label{eqn:App:C_phi}
\mb{C}_{\bs{\phi}} \triangleq \mb{S}_0\bs{\phi} - \mb{C}_0, \quad
\bs{\Gamma}_a = \mb{B}'_a - \mb{A}'_a \mb{C}_{\bs{\phi}}
\end{equation}
where the letter $a\,\in\{\theta,\delta,q_s,\theta_s\}$.
\par By substituting Eq.~(\ref{eqn:App:dA}) and Eq.~(\ref{eqn:App:dB}) into Eq.~(\ref{eqn:App:d_eqn_phi}), and by using the definitions of $\mb{C}_{\bs{\phi}}$ and $\bs{\Gamma}_a$, we have:
\begin{equation}\label{eqn:App:d_eqn_phi_rewritten}
\begin{array}{ll}
&(\mb{A}\mb{S}_0) \;
\left[\begin{array}{c}
\rmd \theta_s \\
\rmd \theta_{\varepsilon}
\end{array}
\right]
-
\left[\begin{array}{c;{2pt/2pt}c}
\bs{\Gamma}_{\theta_s} & \mb{0}
\end{array}\right] \;
\left[\begin{array}{c}
\rmd \theta_s \\
\rmd \theta_{\varepsilon}
\end{array}
\right] = \vspace{3mm}\\
&\bs{\Gamma}_\theta \, \rmd \theta +
\bs{\Gamma}_\delta \, \rmd \delta +
\bs{\Gamma}_{q_s} \, \rmd q_s +
\sum\nolimits_{i}^{n_k} \mb{B}'_{k_{\lambda_i}} \rmd k_{\lambda_i}
\end{array}
\end{equation}
%
This equation shows the full differentiation of Eq. (\ref{eqn:Kin:d_phi_full}) and all the Jacobians can be obtained directly by their definitions, i.e., the expressions of $\left\{\tfrac{\partial \bs{\phi}}{\partial \theta}, \tfrac{\partial \bs{\phi}}{\partial \delta}, \tfrac{\partial \bs{\phi}}{\partial q_s}, \tfrac{\partial \bs{\phi}}{\partial k_{\lambda_i}} \in\realfield{2\times1}\right\}$ may be written as:
\begin{align}
	& \dfrac{\partial \bs{\phi}}{\partial \theta} =
	\left(\mb{A}\,\mb{S}_0 - \bs{\Gamma}_{\theta_s}\mb{S}_1\right)^{-1}\;\bs{\Gamma}_\theta \label{eqn:App:d_phi_d_theta}\\
	& \dfrac{\partial \bs{\phi}}{\partial \delta} =
	\left(\mb{A}\,\mb{S}_0 - \bs{\Gamma}_{\theta_s}\mb{S}_1\right)^{-1}\;\bs{\Gamma}_\delta  \label{eqn:App:d_phi_d_delta}\\
	&  \dfrac{\partial \bs{\phi}}{\partial q_s}  =
	\left(\mb{A}\,\mb{S}_0 - \bs{\Gamma}_{\theta_s}\mb{S}_1\right)^{-1}\; \bs{\Gamma}_{q_s} \label{eqn:App:d_phi_d_q_s}\\
	& \dfrac{\partial \bs{\phi}}{\partial k_{\lambda_i}} =
	\left(\mb{A}\,\mb{S}_0 - \bs{\Gamma}_{\theta_s}\mb{S}_1\right)^{-1}\;\mb{B}'_{k_{\lambda_i}}
	\label{eqn:App:d_phi_d_k_lambda}
\end{align}
where $\mb{S}_1=[1,0]$ is just a selection matrix. 

\section{Derivation of The Jacobian Partitions for the Multi-segment case}\label{app:jacobian_partitions}
Equations (\ref{eqn:Jacobian_partitions}), (\ref{eqn:Kin:2seg:J_x_phi}) and (\ref{eqn:Kin:2seg:J_x_delta_vec}) refer to the Jacobian matrix partitions for the two-segment case where the first segment is the inserted subsegment (indicated vby superscript $s$) and the second segment is the empty subsegment (indicated by subscript $\varepsilon$).  The Jacobian partitions represent the effects of differentials on $\theta_i$ and $\delta_i$ that contribute to the end-effector's translational and rotational differential, labeled by subscripts `$\mb{v}$' and `$\bs{\omega}$', indicating `velocity' and `angular~velocity', respectively. The expressions of \{$\mb{J}_{\mb{v}\theta_i}$, $\mb{J}_{\bs{\omega}\theta_i}$, $\mb{J}_{\mb{v}\delta_i}$, $\mb{J}_{\bs{\omega}\delta_i}$\} are extracted from \cite{Xu2010} as:
\begin{align}\label{eqn:Jacobian_partitions_defs}
	&
	\mb{J}_{\mb{v}\theta_i} =
	D_i
	\begin{bmatrix}
		c_{\delta_i} \;\chi_{a_i}\\
		-s_{\delta_i}  \;\chi_{a_i}\\
		\chi_{b_i}
	\end{bmatrix}, \quad
	\mb{J}_{\bs{\omega}\theta_i} =
	\begin{bmatrix}
		-s_{\delta_i} \\
		-c_{\delta_i} \\
		0
	\end{bmatrix}\\
	&
	\mb{J}_{v\delta_i} =
	D_i
	\begin{bmatrix}
		s_{\delta_i} \;\chi_{c_i}\\
		c_{\delta_i}  \;\chi_{c_i}\\
		0
	\end{bmatrix}, \quad
	\mb{J}_{\omega\delta_i} =
	\begin{bmatrix}
		c_{\delta_i} s_{\theta_i}\\
		-s_{\delta_i} c_{\theta_i}\\
		-1 + s_{\theta_i}
	\end{bmatrix}
\end{align}
Where $c(\cdot)$ and $s(\cdot)$ denote the cosine and sine functions, and $D_i$ represents the length of the subsegment. For the inserted subsegment, $D_s=q_s$; and for the empty subsegment, $D_r=L-q_s$. In addition, the following shorthanded notations are used:
\begin{align}
	&\chi_{a_i} = \dfrac{(\theta_i - \theta_0)c_{\theta_i} - s_{\theta_i} + 1}{(\theta_i-\theta_0)^2}\\
	& \chi_{b_i} = \dfrac{(\theta_i - \theta_0)c_{\theta_i} + c_{\theta_i}}{(\theta_i-\theta_0)^2}
	\quad , \quad
	\chi_{c_i} = \dfrac{s_{\theta_i} - 1}{\theta_0 - \theta_i}\label{eqn:Kin:2seg:chi_c}
\end{align}
%
\balance

\end{document}